\documentclass[preprints,article,accept,pdftex,moreauthors]{Definitions/mdpi} 

\firstpage{1} 
\pubvolume{1}
\issuenum{1}
\articlenumber{0}
\pubyear{2023}
\copyrightyear{2023}
\datereceived{ } 
\daterevised{ } 
\dateaccepted{ } 
\datepublished{ } 
\hreflink{https://doi.org/} 

\usepackage{amsfonts}       
\usepackage{nicefrac}       
\usepackage{lipsum}
\usepackage{import}
\usepackage{color}
\usepackage{layout}
\usepackage{mathtools}
\usepackage{systeme}
\usepackage{subcaption}
\usepackage{multirow}
\usepackage{array}
\allowdisplaybreaks

\Title{\MakeUppercase{Autoencoder-Based Visual Anomaly Localization for Manufacturing Quality Control}}

\TitleCitation{Title}

\Author{Devang Mehta $^{1}$\orcidA{} and Noah Klarmann $^{1}$*\orcidB{}} 

\AuthorNames{Devang Mehta and Noah Klarmann}

\AuthorCitation{Mehta, D.; Klarmann, N.}

\address{%
	$^{1}$ \quad Faculty of Management and Engineering, Rosenheim Technical University of Applied Sciences, Hochschulstraße 1, 83024 Rosenheim, Germany}
	
\corres{Correspondence: noah.klarmann@th-rosenheim.de}

\abstract{Manufacturing industries require efficient and voluminous production of high-quality finished goods. In the context of Industry 4.0, visual anomaly detection poses an optimistic solution for automatically controlled product quality with high precision. In general, automation based on computer vision is a promising solution to prevent bottlenecks at the product quality checkpoint. We considered recent advancements in machine learning to improve visual defect localization, but challenges persist in obtaining a balanced feature set and database of the wide variety of defects occurring in the production line. Hence, this paper proposes a defect localizing autoencoder with unsupervised class selection by clustering with k-means the features extracted from a pre-trained VGG16 network. Moreover, the selected classes of defects are augmented with natural wild textures to simulate artificial defects. The study demonstrates the effectiveness of the defect localizing autoencoder with unsupervised class selection for improving defect detection in manufacturing industries. The proposed methodology shows promising results with precise and accurate localization of quality defects on melamine-faced boards for the furniture industry. Incorporating artificial defects into the training data shows significant potential for practical implementation in real-world quality control scenarios.}

\begin{document}

\section{\MakeUppercase{Introduction}}
Artificial Intelligence (AI) promises to be a revolutionary force in the 21st century. It has gained significant attention across various sectors, with extensive research, development, and production of AI-driven products and services. The widespread adoption, ease of use, and flexibility of AI technologies have propelled its evolution. This research aims to contribute this revolutionary wave to visual inspection in furniture manufacturing.

In the manufacturing industry, the visual inspection of products plays a crucial role at different stages of the production process; ensuring the quality of the final product is essential to meet aesthetic and functional requirements. At the University of Applied Sciences Rosenheim, the \href{https://www.th-rosenheim.de/en/die-hochschule/labore/proto-lab}{proto\_lab}\footnote{\href{https://www.th-rosenheim.de/en/die-hochschule/labore/proto-lab}{https://www.th-rosenheim.de/en/die-hochschule/labore/proto-lab}} laboratory is an innovative Industry 4.0 platform that produces furniture with state-of-the-art machinery in a fully digitalized way, posing an ideal ecosystem for applying AI use cases. The goal of applying data-based methodologies is to establish a high-quality, efficient, intelligent system to improve the production cycle holistically. In this context, an integrated autonomous system, particularly computer vision (CV) systems, has proven to be an overly promising way. Detecting product flaws is denoted as anomaly detection, alternatively also known as artifact, novelty, or outlier detection.

\citet{li2023research} presents an overview on machine vision applications in furniture manufacturing from 2011 to 2022. Many/most studies rely on classical methods to perform quality checks. These traditional methodologies are well-established, straightforward, and computationally optimized but usually show limitations. Limitations may occur due to variations in the environment, specific manual step-wise feature engineering, heterogeneous images regarding size and quality, and complexity of the image data. While widely utilized for elementary implementations and tasks, classical methods cannot keep pace with technological advancements in camera sensor resolution, optics, and Deep Learning (DL) techniques \cite{omahony2020advances}. DL methods leverage large amounts of data, requiring minimal expertise and automatic fine-tuning. DL demonstrates flexibility to adapt to different domains and datasets, even when the relevant data is limited – e.g., by making use of transfer learning \cite{weiss2016survey}. The dataset's images are sharp and consistent in feature space in the scope of the problem. Classical CV methods often require algorithmic computation for each type of feature or class, making them expensive to implement. Anomaly localization, however, can be efficiently scaled up by leveraging the advantages of neural networks. Several DL methodologies have shown remarkable predictive performance. In some cases, hybrid approaches might be the best choice when combining traditional CV methods and DL techniques. 

As manufacturing systems excel in optimization, productivity, and efficiency, the number of products having defects reduces enormously. Consequently, current research and development efforts are in unsupervised anomaly localization, with generative deep neural networks gaining significant prominence. Initially, a popular approach was to train models using non-anomalous classes and predict classes of anomalous and non-anomalous instances. However, this methodology requires additional information for generating or reconstructing non-anomalous images. Modern challenges include accurately localizing anomalies within a low-variance feature set in an image dataset. Achieving an intelligent, thorough, fast, robust, and reliable CV system necessitates the integration of cutting-edge technologies such as deep learning and 3D point cloud analysis in some cases where depth information is required \cite{li2023research}.

This paper presents a hybrid approach for localizing surface defects on melamine-faced boards. In contrast to directly focusing the camera on specific dimensions of boards, our image dataset consists of high-resolution images captured with a camera having a fixed field of view, allowing inspection of boards of various proportions. Achieving such a model is done by slicing the high-resolution image and selecting classes of interest from an imbalanced dataset with feature extraction and k-means clustering in order to consider the minute variation in the frequency of the features, and finally, simulating anomalies for the autoencoder model to predict all artifacts.

The subsequent sections delve deeper into the literature, methodology, results, and implications of this research, providing an understanding of the approaches utilized in localizing anomalies in furniture manufacturing.
\section{\MakeUppercase{Technical background}}
In the following, we present an overview of the core methodologies and principles that underpin our research. For readers well-versed in these methodologies, this section may be familiar and can skip to the next section without losing continuity in the paper. However, for those less acquainted with these concepts, this section serves as a short introduction, illuminating the methods and tools utilized throughout our work.

\subsection{\MakeUppercase{K-means clustering}}
\label{sub:sec:k-means}
The unsupervised K-means algorithm, as introduced by \citet{hartigan1979algorithm}, divides \textit{M} points in \textit{N} dimensions into \textit{K} clusters in a way that within every cluster, the sum of squares are minimized. Typically, K-means executes as a preprocessing step before the start of the main algorithm. Generally, clustering has a somewhat simple implementation, and the algorithm is guaranteed to converge. It easily scales up to adapt to new data with mini-batches to save computation time, cluster merging when the new data clusters based on existing centroids are close, centroid initialization by utilizing existing centroids, online k-means for continuous data or incremental Principal Component Analysis (PCA) when the dimensionality of the data is very high. The k-means algorithm degrades in performance when the data contains several outliers, resulting in incorrect clustering. Because the k-means algorithm works by minimizing the sum of the squared distance between the data point and the centroid, some data would be clustered tightly in high-density regions, while other regions would have data spread farther apart. Hence, it is worth noting that the k-means algorithm must be generalized for robust performance when the data is of varying density.

\subsection{\MakeUppercase{Segmentation}}
\label{sub:sec:segmentation}
\begin{figure}[H]
	\centering
	\begin{subfigure}{0.34\textwidth}
		\centering
		\includegraphics[width=\linewidth]{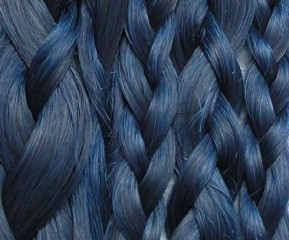}
		\caption{}
		\label{fig:sub:f0}
	\end{subfigure}
	\begin{subfigure}{0.34\textwidth}
		\centering
		\includegraphics[width=\linewidth]{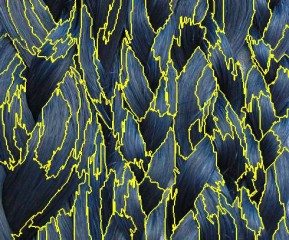}
		\caption{}
		\label{fig:sub:f1}
	\end{subfigure}
	\caption{DTD (a) image of braid, and (b) its Felzenszwalb segmented boundaries.}
	\label{fig:segmentation}
\end{figure}

Image segmentation aims to simplify and represent an image by dividing it into multiple regions denoted as segments. Each segment represents pixels that share the same features. This process helps to analyze the image data more meaningfully in a spatial sense. Unlike clustering, segmentation also considers boundaries and structures. Various segmentation algorithms use specific conditions to segment the image. \citet{felzenszwalb2004efficient} introduced a fast algorithm that generates segments based on a boundary that separates the regions. By applying Felzenszwalb’s segmentation with $scale = 25$, $sigma = 1$, and $min\_size = 500$ to an image of braid from DTD Figure~\ref{fig:sub:f0}, the algorithm can find segments as shown in Figure~\ref{fig:sub:f1}.

\subsection{\MakeUppercase{SSIM}}
\label{sub:sec:ssim}
\begin{figure}[H]
	\centering
	\begin{subfigure}{0.34\textwidth}
		\centering
		\includegraphics[width=0.99\linewidth]{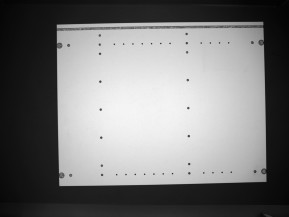}
		\caption{}
		\label{fig:sub:ssim0}
	\end{subfigure}%
	\begin{subfigure}{0.34\textwidth}
		\centering
		\includegraphics[width=0.99\linewidth]{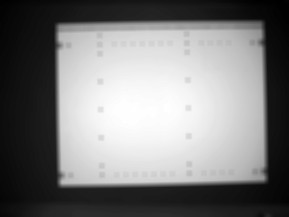}
		\caption{}
		\label{fig:sub:ssim1}
	\end{subfigure}
	\caption{Comparing melamine-faced board sample image (a) actual vs. (b) blurred image.}
	\label{fig:ssim}
\end{figure}

Structural Similarity (SSIM), introduced by \citet{wang2004image}, is a commonly used algorithm to determine the similarity or difference between two images. The algorithm is designed specifically for grayscale images, considering their inherent properties such as luminance, contrast, and structure. It adopts mechanisms of human vision for effectively identifying structural information. The SSIM score ranges from zero to one, while unity indicates perfect similarity and zero indicates complete dissimilarity. For example, the SSIM score between the original image Figure~\ref{fig:sub:ssim0} and its blurred version Figure~\ref{fig:sub:ssim1} with a kernel of $(99, 99)$ is $0.9646$.

\subsection{\MakeUppercase{Autoencoders}}
\label{sub:sec:ae}
The generative model autoencoder is a class of unsupervised learning algorithms in which the output shape is the same as the input shape \cite{vasilev2019python}. Such a model allows the network to learn basic representations in a better way when compared to raw, unprocessed data, thereby learning features and patterns while ignoring noise. The network has an encoder that maps the information to a latent representation. Following it is decoding the latent space to reconstruct the original data. The model optimizes by minimizing the MSE loss between the target and the reconstructed image. Usually, the convolutional autoencoders (CAEs) are not very promising for localizing anomalies; using a de-noising autoencoder (DAE) with an SSIM-based loss typically increases the performance of CAE.

\subsection{\MakeUppercase{Generative Adversarial Networks}}
\label{sub:sec:gan}
Generative Adversarial Networks (GANs), as described by \citet{goodfellow2014generative}, are generative models based on game theory, where the players (the generator and the discriminator) each try to beat the other through strategic adjustments. In the context of image data, the model works competitively, where a generator creates images from random noise that resemble real images, and the discriminator distinguishes between the real and the generated images based on a probability score. In short, GANs learn the probability distribution of the data to generate synthetic data. GAN-based methods fail where the discriminator gets stuck in a local minimum, and the generators produce a particular output repeatedly, making it difficult to reliably reconstruct anomaly-free images, especially textures \cite{hida2021smart}.

\subsection{\MakeUppercase{U-net}}
\label{sub:sec:unet}
U-net, as introduced by \citet{ronneberger2015u}, was initially designed for segmentation in the field of medical imaging. It depends on data augmentation and can be trained even with a couple of images. The main difference between autoencoder and an u-net architecture is the implementation of skip connections. Skip connections help the network maintain high-resolution features lost during downsampling.

\subsection{\MakeUppercase{Feature extraction with pre-trained CNN}}
\label{sub:sec:pretrainedcnn}
\begin{figure}[H]
	\centering
	\def\svgwidth{1.1\textwidth}
	\resizebox{0.975\textwidth}{!}{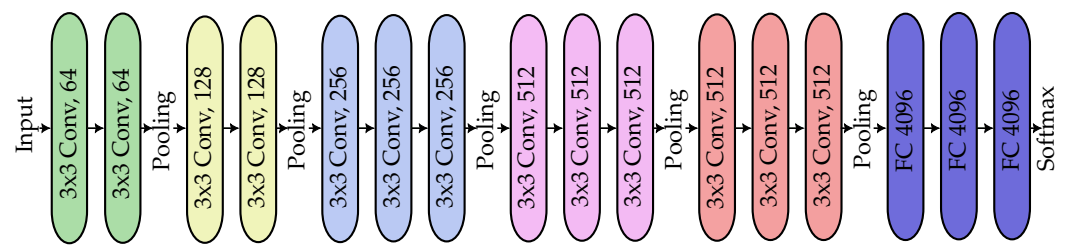}
	\caption{The VGG16 model architecture.}
	\label{fig:featureextraction}
\end{figure}

A deep neural network can extract feature descriptors for a completely different dataset. This approach is known as feature extraction. Figure~\ref{fig:featureextraction} depicts the architecture of the popular VGG16 \cite{simonyan2014very} deep neural network architecture. The first part of the fully connected layer of the model provides the features of the input image. Simple clustering or classification utilizes these feature vectors from multiple dataset images as post-processing.

\section{\MakeUppercase{Related Work}}
Anomaly detection tackled with AI methodologies broadly categorizes into supervised, semi-supervised, and unsupervised models. The lack of anomaly ground truth values in an image dataset is the foremost cause for researchers to employ unsupervised anomaly localization models. Due to the lack of data on anomalous classes, modern research and development in unsupervised anomaly localization relies on reconstructive or generative deep neural networks. A popular method that follows \citet{goodfellow2016deep} is to train the model with anomaly-free classes only, and the difference between the reconstruction and the input data would localize the anomalies. The methodology suffers from insufficient information on generating or reconstructing an image of a non-anomalous class.

Based on the criteria of the type of methodology incorporated to try to localize anomalies, we have two subsections that provide a better understanding of the existing research. The studies follow two approaches, one where the input data is encoded in some representation and then fed for training, and the other where the focus is mainly on identifying the difference with the reconstruction.

\subsection{\MakeUppercase{Encoding-based anomaly localization}}
The presence of knots in wood plays a crucial role in assessing the eventual quality/strength of the end product. \citet{kamal2017wood} introduces a unique technique, employing feed-forward back-propagation neural networks with Laws Texture Energy Measures (LTEM) \cite{laws1979texture} as input parameters. This innovative approach aims to predict knot defects in wood through a supervised classification model. While the model performs well for multi-class classification, it struggles with generalization when dealing with unfamiliar defects. \citet{nakanishi2021anomaly} proposes an alternative approach using autoencoders and a weighted frequency domain loss to identify various wood anomalies effectively. The study reveals that the weighted frequency domain loss significantly improves the autoencoder’s ability to detect anomalies by emphasizing specific frequency components. However, we need further investigation to assess its effectiveness on real-time datasets and, as mentioned by the authors, on high-frequency components.

\subsection{\MakeUppercase{Reconstruction-based anomaly localization}}
The widely recognized MVTec AD dataset, established by \citet{bergmann2019mvtec}, serves as the standard benchmark for unsupervised anomaly detection and localization. Many researchers utilize the autoencoder architecture for feature learning on this dataset and employ image inpainting techniques to enhance the robustness of generalized predictions. DRAEM, as introduced by \citet{zavrtanik2021draem}, is a notable model that is a reconstructive-discriminative sub-network trained on the MVTec AD dataset specifically for visual surface anomaly detection. DRAEM incorporates a Perlin noise generator \cite{perlin1985image} to produce random shapes of artificial anomalies, while the shape content derives from randomly augmented DTD \cite{cimpoi2014describing} images. DRAEM outperforms several other methods, achieving impressive mean detection/localization; e.g., an AUROC score of $0.98$ and $0.973$, respectively, could be achieved. It improves the accuracy of anomaly localization and achieves nearly fully supervised results on surface defect datasets. 

Another novel model named CS-ResNet, proposed by \citet{zhang2021cs}, is introduced for PCB cosmetic defect detection using convolutional neural networks. It addresses issues related to unbalanced class distribution and misclassification cost by incorporating a cost-sensitive adjustment layer in the standard ResNet \cite{he2016deep}. This modification results in higher accuracy and lower misclassification cost compared to Auto-VRS \cite{deng2018building}.

DeRA, as introduced by \citet{hida2021smart} and akin to DRAEM, is an unsupervised anomaly detection method tested on the MVTec AD dataset. It leverages Felzenszwalb’s graph-based segmentation method \cite{felzenszwalb2004efficient} on segments of the DTD \cite{cimpoi2014describing} dataset. These segments are superimposed on non-anomalous images with random transparency, creating a more diverse and complex anomalous dataset. The DeRA neural network combines U-net \cite{ronneberger2015u} and autoencoder architectures. The performance of DeRA is notably enhanced by incorporating the Neural Style Transfer (NST) loss \cite{jing2019neural} using a pre-trained VGG19 \cite{simonyan2014very} network as a discriminator. This loss function, trained on the ImageNet dataset \cite{deng2009imagenet}, aids in improving the anomaly detection results. DeRA achieves a pixel-wise mean AUROC of $0.97$, surpassing methodologies described by \citet{bergmann2019mvtec}. However, this method is limited to grayscale images and performs poorly on transparent artifacts.

In contrast to DeRA, \citet{schlueter2022natural} propose a self-supervision task called Natural Synthetic Anomalies (NSA) using Poisson image editing \cite{perez2003poisson} to generate a wide range of realistic synthetic anomalies for anomaly detection and localization. The NSA architecture, a ResNet-based encoder-decoder, achieves mean image-level and pixel-level AUROC scores of $0.972$ and $0.963$ on the MVTec AD dataset. While outperforming particular methods that do not use additional data, NSA lacks robustness when dealing with minute anomalies.

To compare different autoencoder models for real-time anomaly detection, \citet{mujkic2022anomaly} evaluated the following three models: Denoising Autoencoder (DAE) \cite{vincent2008extracting}, Semi-Supervised Autoencoder (SSAE), and Variational Autoencoder (VQ-VAE) \cite{van2017neural}, against the baseline YOLOv5 \cite{jocher2022ultralytics}. Although YOLOv5 slightly outperformed SSAE in the AUROC score ($0.945$ vs. $0.8849$), SSAE demonstrated better performance in critical cases.

Lastly, \citet{huang2022self} introduced a self-supervised masking (SSM) method for anomaly localization. They use random masking to augment each image, creating a diverse set of training triplets \cite{hoffer2015deep}, enabling the autoencoder to reconstruct masks of various shapes and sizes during training. For inference, a progressive mask refinement method gradually reveals non-anomalous regions and eventually localizes anomalies. On the MVTec AD dataset, the SSM achieves a mean AUROC score of $0.92$.

Our research paper leverages the autoencoder architecture, incorporating the autoencoder architecture, an artificial anomaly overlay, and a loss function akin to DeRA. Unlike conventional approaches, our unique pipeline accommodates diverse features within our training dataset. Given the real-world limitations of training high-resolution images directly, we adopt the sliding window technique to capture and comprehend image features effectively. Implementing the sliding window approach, we address memory constraints and augment the richness of our training data. Such an implementation additionally allows our model to be versatile for all dimensions of the boards without an additional camera system. Different from CS-ResNet, we integrate k-means clustering to identify and prioritize non-frequent image features that could be overshadowed by dominant classes, refining the focus of our model and rectifying class imbalances. The resultant balanced training data propels the overall efficacy of our approach. Similar to the methodology in DRAEM, we generate artificial anomalies akin to DeRA using Felzenszwalb’s method. Our workflow stands for an innovative and pragmatic solution for advancing anomaly detection within complex image datasets.

\section{\MakeUppercase{Methodology}}
\begin{figure}
	\centering
	\def\svgwidth{1.4\textwidth}
	\resizebox{\textwidth}{!}{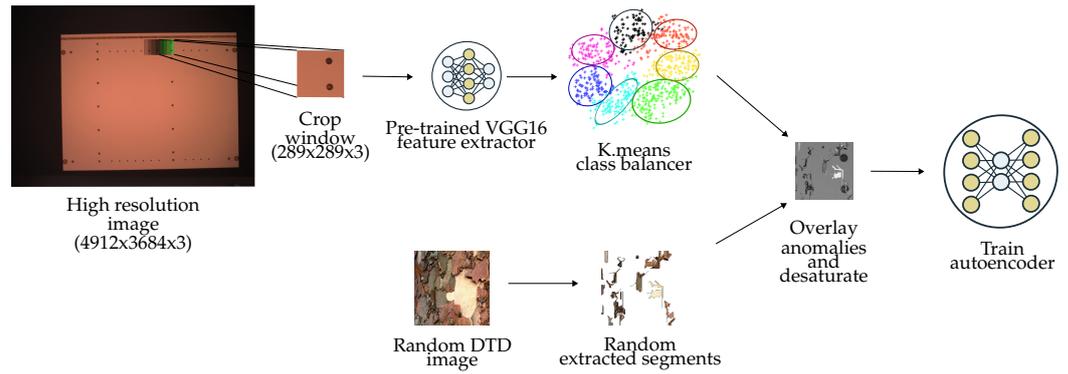}
	\caption{Unsupervised defect localizing training pipeline.}
	\label{fig:pipeline}
\end{figure}

This paper aims to provide a novel solution for localizing artifacts on a melamine-faced board surface subject on the basis of a highly imbalanced data set. The methodology deals with the frequently arising problem of insufficient data featuring anomalies and requiring a high-resolution image for anomaly identification. The first step in the approach is to capture high-resolution images of the melamine-faced boards. Moreover, we employ the sliding window technique to slide a small window to slice the high-resolution images into small crops, as detailed in Subsections \ref{sub:sec:imaging} and \ref{sub:sec:imageanalysis}. The features of the sliced images are extracted with the help of a pre-trained VGG16 and clustered into groups with the k-means algorithm, as described in Subsection \ref{sub:sec:classselection}. As a separate process, Subsection \ref{sub:sec:augment} explains the artificial anomaly generation by extracting segments from the images of DTD. Finally, the artificial anomalies are dynamically and randomly overlaid on the fly for training. Figure~\ref{fig:pipeline} represents the pipeline of the method, while the following subsections detail the elements.

\subsection{\MakeUppercase{Imaging system and dataset gathering}}
\label{sub:sec:imaging}
\begin{figure}[H]
	\centering
	\begin{subfigure}{0.34\textwidth}
		\centering
		\includegraphics[width=\linewidth]{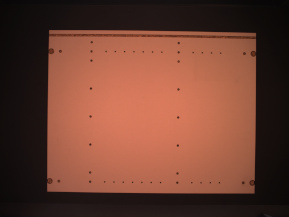}
		\caption{Lateral orientation}
		\label{fig:sub:prototype_lateral}
	\end{subfigure}
	\begin{subfigure}{0.34\textwidth}
		\centering
		\includegraphics[width=\linewidth]{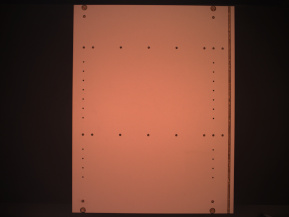}
		\caption{Transverse orientation}
		\label{fig:sub:prototype_transverse}
	\end{subfigure}
	\caption{Prototype setup training image.}
	\label{fig:prototype_image}
\end{figure}

The demonstrator is operated at controlled conditions, featuring an enclosed lightproof setup utilizing a precise artificial light source, camera position, and focus. A graphical user interface (GUI) running on a Raspberry Pi 4 \cite{web2023raspberry} empowers the system to initiate image capturing when a product sample takes its position beneath the anticipatory camera lens. Our data collection venture encompassed a multitude of captures, each revealing the melamine-faced board’s distinctive facets through diverse locations and orientations within the enclosure. The resulting image dataset showcases a lateral viewpoint (Figure~\ref{fig:sub:prototype_lateral}) alongside a transverse (Figure~\ref{fig:sub:prototype_transverse}). The image dataset presents invaluable insights, culminating in 348 images, for our analysis and model development pursuits.

\subsection{\MakeUppercase{Image dataset analysis}}
\label{sub:sec:imageanalysis}
The demonstrator captures images with a resolution of $(4912, 3684)$. Each melamine-faced board has a barcode label glued on its surface. However, the label feature space resembles the melamine-faced board's surface feature space, presenting a critical issue. Hence, the barcode labels require manual removal beforehand. Seven distinct classes meticulously considered in the images are as follows: (a) background, (b) surface, (c) drilling holes, (d) edges, (e) corners, (f) slots, and (g) combinations of edges, holes, and slots. The board carefully selected for this endeavor is Figure~\ref{fig:prototype_image} because it included the most diverse classes, with each board exhibiting unique features. 

\begin{figure}[H]
	\centering
	\begin{subfigure}{0.118\textwidth}
		\centering
		\includegraphics[width=\linewidth]{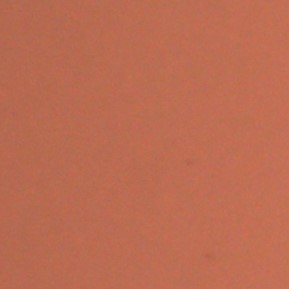}
		\caption{}
		\label{fig:sub:p_board}
	\end{subfigure}
	\begin{subfigure}{0.118\textwidth}
		\centering
		\includegraphics[width=\linewidth]{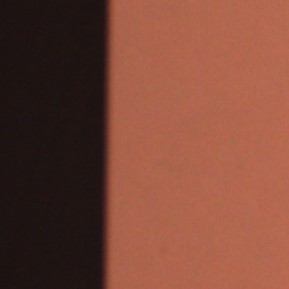}
		\caption{}
		\label{fig:sub:p_edge}
	\end{subfigure}
	\begin{subfigure}{0.118\textwidth}
		\centering
		\includegraphics[width=\linewidth]{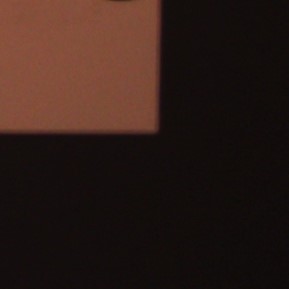}
		\caption{}
		\label{fig:sub:p_corner}
	\end{subfigure}
	\begin{subfigure}{0.118\textwidth}
		\centering
		\includegraphics[width=\linewidth]{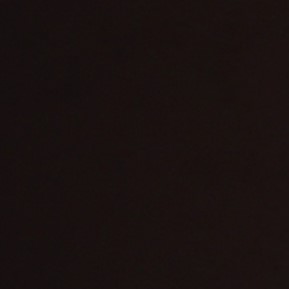}
		\caption{}
		\label{fig:sub:p_background}
	\end{subfigure}
	\begin{subfigure}{0.118\textwidth}
		\centering
		\includegraphics[width=\linewidth]{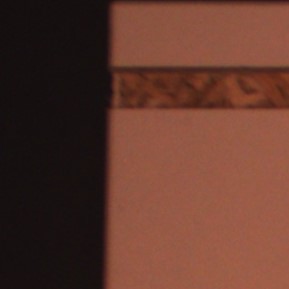}
		\caption{}
		\label{fig:sub:p_edgegroove}
	\end{subfigure}
	\begin{subfigure}{0.118\textwidth}
		\centering
		\includegraphics[width=\linewidth]{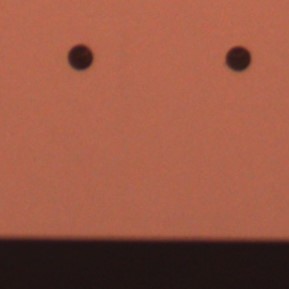}
		\caption{}
		\label{fig:sub:p_holeedege}
	\end{subfigure}
	\begin{subfigure}{0.118\textwidth}
		\centering
		\includegraphics[width=\linewidth]{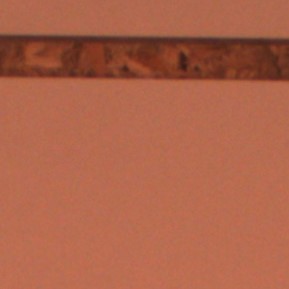}
		\caption{}
		\label{fig:sub:p_groove}
	\end{subfigure}
	\caption{Sample windows of furniture finishing (a) Surface, (b)  Edge, (c)  Corner, (d) Background, (e) Edge \& groove, (f) Hole \& edge, (g) Groove}
	\label{fig:p_windows}
\end{figure}

The challenge arose in training the network directly on such high-resolution images. Hence, the crop windows of the image, Figure~\ref{fig:p_windows}, each of resolution $(289, 289)$, are generated with a precise stride of $(67, 97)$ and zero padding. This approach resulted in 2'520 cropped tiles derived from a single high-resolution image and 876'960 tiles extracted from all images (348 images in total).

The image dataset demonstrated exceptional consistency in the feature space, featuring only minute variance. This consistency instrumentally overcame the inherent challenges of lens distortion, where features further away from the camera are captured with a bulge, elongation, or stretch (barrel distortion) without camera calibration.

\subsection{\MakeUppercase{Unsupervised class selection}}
\label{sub:sec:classselection}
The prototype image dataset features noticeable class imbalances as surface and background tiles dominate the remaining classes. To mitigate this issue, we adopt an approach of two steps: Firstly, we extract feature vectors for each cropped tile using the pre-trained VGG16 model with ImageNet1k\_v1 weights. This step significantly contributes to resolving the class imbalance by enhancing the representation of each class. Next, we utilize the unsupervised k-means algorithm to cluster the obtained feature vectors into seven distinct clusters. In doing so, we identify inherent patterns and group samples regarding their similarity. Moreover, we deliberately drop the clusters with the highest frequency, corresponding to the surface class, and the second-highest frequency, associated with the background class. By discarding these clusters, we ensure that only the most relevant and informative ones are retained, thus facilitating a more balanced and meaningful representation of the dataset.

\subsection{\MakeUppercase{Image dataset augmentation}}
\label{sub:sec:augment}
\begin{figure}[H]
	\centering
	\includegraphics[width=0.5\linewidth]{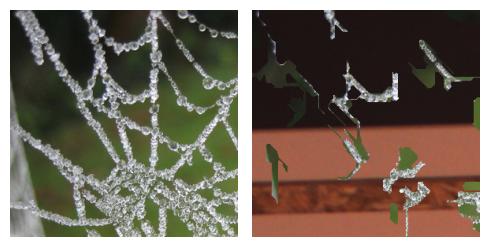}
	\caption{DTD (left) \& artificial anomaly overlaid on crop window of prototype dataset (right)}
	\label{fig:artificialanomaly}
\end{figure}

Data augmentation in computer vision is the process employed to increase the diversity of a training dataset by applying diverse transformations such as rotation, flipping, scaling, and cropping to the original images. This technique helps to enhance the ability of the model to generalize by exposing it to different variations of the data, thereby improving performance and reducing overfitting. We achieve data augmentation by adopting the extraction of segments from the Describable Textures Dataset (DTD) \cite{cimpoi2014describing}. The DTD is a collection of natural patterns and textures, serving as a foundation for developing better methods to recognize and understand texture attributes in images. Notably, the DTD contains image resolutions higher than the cropped-out tiles resulting from the sliding window technique. We employ Felzenswalb’s segmentation algorithm to extract segments from the DTD with specific values for the algorithm parameters, as detailed in Table~\ref{table:felzenswalb}. Next, the resulting image segments are randomly selected and superimposed onto the tile, as demonstrated in Figure~\ref{fig:artificialanomaly}. This approach enables the generation of diverse artificial anomaly variations, significantly contributing to the overall performance of the anomaly detection model and increasing its robustness to detect anomalous patterns in various real-world scenarios. This augmentation strategy is instrumental in improving the model’s generalization capabilities and practical applicability in real-world anomaly detection tasks. We additionally employ data augmentation techniques, as detailed in  Table~\ref{table:augmentation}. During the training process, these augmentations are randomly applied to each image, enriching the diversity of the data and enhancing the model’s ability to generalize effectively.
\begin{minipage}{\linewidth}
	\centering
	\begin{minipage}{0.4\linewidth}
		\centering
		\begin{table}[H]
			\caption{Felzenswalb's parameter values}
			\label{table:felzenswalb}
			\newcolumntype{C}{>{\centering\arraybackslash}X}
			\begin{tabularx}{\textwidth}{CC}
				\toprule
				\textbf{Parameter} & \textbf{Value} \\
				\midrule
				scale     & 2   \\
				sigma     & 5   \\
				min\_size & 100 \\
				\bottomrule
			\end{tabularx}
		\end{table}
		\vfill
	\end{minipage}
	\hspace{3em}
	\begin{minipage}{0.4\linewidth}
		\centering
		\begin{table}[H]
			\caption{Augmentation values}
			\label{table:augmentation}
			\newcolumntype{C}{>{\centering\arraybackslash}X}
			\begin{tabularx}{\textwidth}{CC}
				\toprule
				\textbf{Augmentation}     & \textbf{Value}             \\
				\midrule
				horizontal flip  & \multirow{2}{*}{probability = 0.5} \\
				vertical flip    &  \\
				brightness range & [0.98, 1.5]       \\
				contrast range   & [1, 1.2]          \\
				\bottomrule
			\end{tabularx}
		\end{table}
	\end{minipage}
\end{minipage}

\subsection{\MakeUppercase{Network architecture}}
\begin{figure}[H]
	\centering
	\def\svgwidth{1.2\textwidth}
	\resizebox{0.95\textwidth}{!}{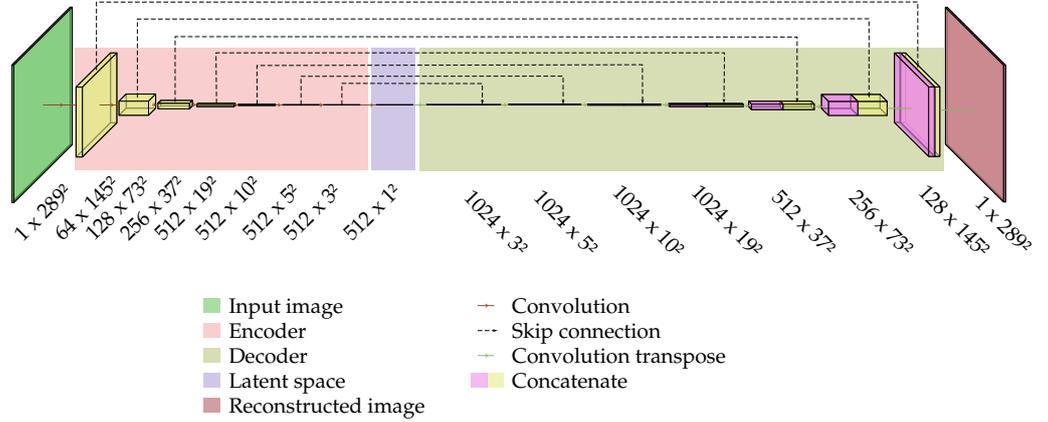}
	\caption{Autoencoder architecture}
	\label{fig:ae_architecture}
\end{figure}

Inspired by the work of \citet{hida2021smart}, we propose the network architecture, as depicted in \ref{fig:ae_architecture}, with the principal objective of encoding an input image of resolution $289 \text{ x } 289$ into a compact latent representation of size $512 \text{ x }  1 \text{ x } 1$ using the encoder. After the encoding, the decoder reconstructs the original input image. As indicated in \ref{fig:ae_architecture}, the autoencoder features skip connections that bypass the middle part of the network. In general, skip connections enable the network to more easily retain finer details from the input, thus facilitating the learning of a more accurate reconstruction of the original data. By bypassing specific layers, skip connections can also alleviate the vanishing gradient problem, which is particularly beneficial when training deeper autoencoders.

\subsection{\MakeUppercase{Loss Function}}
At the outset, we implemented the Mean Squared Error (MSE) loss function alone, which yielded unsatisfactory results. However, we observed a significant improvement after introducing the Structural Similarity Index (SSIM) metric as an additional term in the loss function. This addition notably enhanced the preservation of intricate image details, resulting in greater accuracy. Given that SSIM is for grayscale images, we adjusted the dataset accordingly, leading us to utilize grayscale images for training. Further increasing overall performance, we integrate an extra MSE loss component tailored for overlay regions. This addition proved crucial in emphasizing precise reconstruction within these critical areas, contributing to enhanced outcomes. The primary responsibility for emphasizing accurate reconstruction in these vital areas, where anomalies exist, lies with the MSE at overlay loss term. Its influence leads to improved results, particularly in overlay regions. We culminate these insights and develop the final version of the loss function (see Equation~\ref{eq:lossfn}), which combines MSE (Equation~\ref{eq:mse}), SSIM (Equation~\ref{eq:ssim}), and MSE at overlay (Equation~\ref{eq:mse_ano}) for training. The individual loss weights in Equation~\ref{eq:lossfn}, $\lambda_{MSE}$, $\lambda_{SSIM}$, and $\lambda_{MSE\_artificial\_anomaly}$ use a value of one. In the last step, capitalizing on the strengths of these three distinct loss metrics, we train the model to achieve excellent accuracy and robustness in addressing artificial anomalies and irregularities such as unexpected variations, inconsistencies, abnormal patterns, noise, errors, or outliers in the data. This comprehensive approach ensured our model's proficiency in handling complex challenges within the dataset.

\begin{align}
	Loss = & \lambda_{MSE}L_{MSE}(Y,\hat{Y}) +\notag\\
	& \lambda_{SSIM}(1 - L_{SSIM}(Y,\hat{Y})) +\notag\\
	& \lambda_{MSE\_artificial\_anomaly}L_{MSE\_artificial\_anomaly}(Y,\hat{Y})
	\label{eq:lossfn}
	\\
	L_{MSE}(Y,\hat{Y}) = & \frac{1}{p}\sum_{i=1}^{p}(Y_i - \hat{Y}_i)
	\label{eq:mse}
	\\
	L_{SSIM}(Y,\hat{Y}) = & \frac{1}{q}\sum_{j=1}^{q}\frac{(2\mu_{Y_{j}}\mu_{\hat{Y}_j} + c_1)(2\sigma_{Y_j\hat{Y}_j} + c_2)}{(\mu_{Y_j}^2 + \mu_{\hat{Y}_j}^2 + c_1)(\sigma_{Y_j}^2 + \sigma_{\hat{Y}_j}^2 + c_2)},
	\label{eq:ssim}
	\\
	& where \text{ } c_1 = 0.01, \text{ } and \text{ } c_2 = 0.03 \notag
	\\
	L_{MSE\_artificial\_anomaly}(Y,\hat{Y}) = & \frac{1}{r}\sum_{k=1}^{r}(Y_{k} - \hat{Y}_{k}),
	\label{eq:mse_ano}
	\\
	& where \text{ } k \text{ } \epsilon \text{ } \{artificial \text{ } anomaly \text{ } pixels\} \notag
\end{align}

\subsection{\MakeUppercase{Training}}
\begin{figure}[H]
	\centering
	\begin{subfigure}{0.475\textwidth}
		\centering
		\def\svgwidth{1.25\textwidth}
		\resizebox{\textwidth}{!}{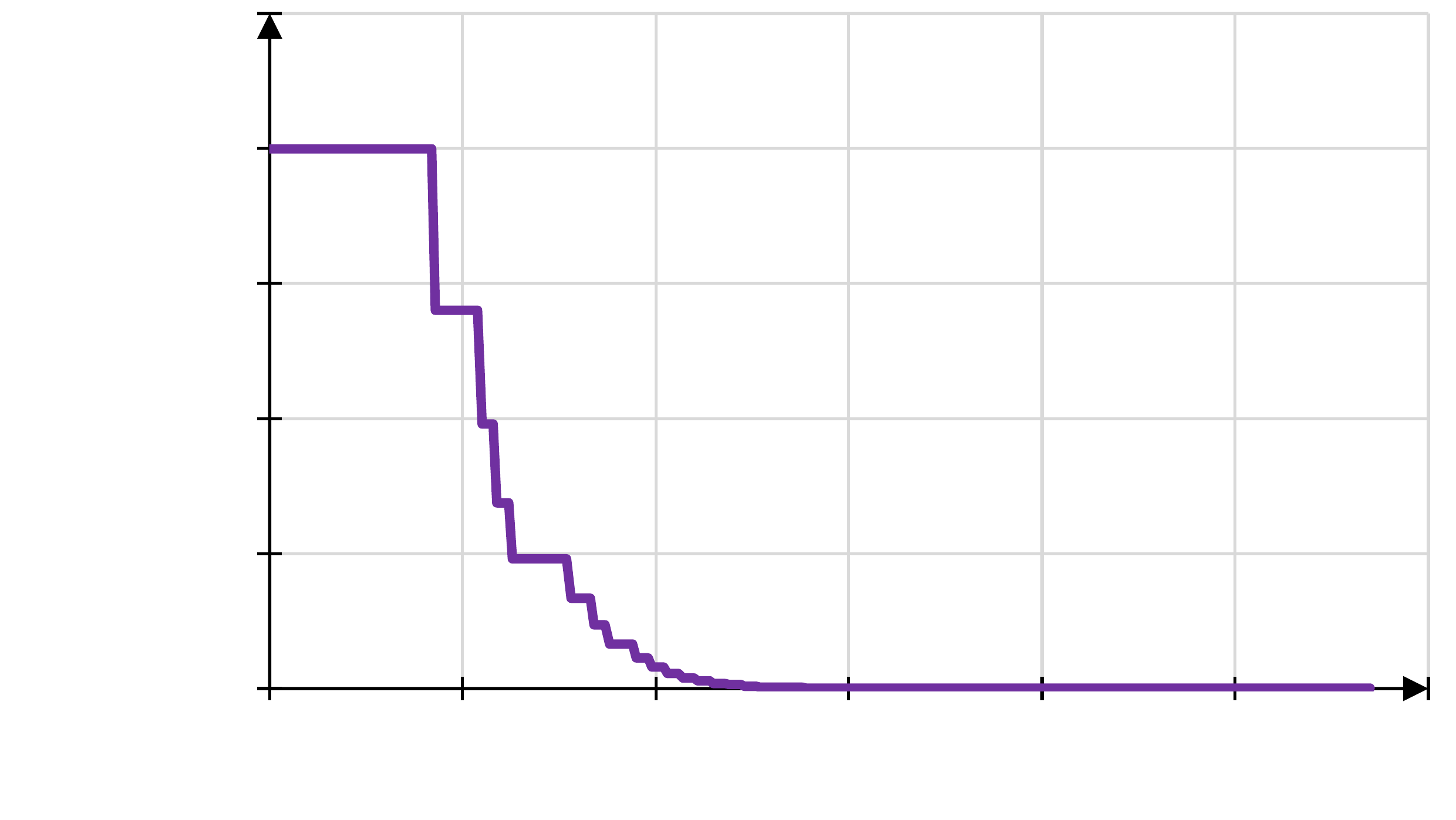}
		\caption{Learning rate schedule}
		\label{sub:fig:lr}
	\end{subfigure}
	\hspace{1em}
	\begin{subfigure}{0.475\textwidth}
		\centering
		\def\svgwidth{1.25\textwidth}
		\resizebox{\textwidth}{!}{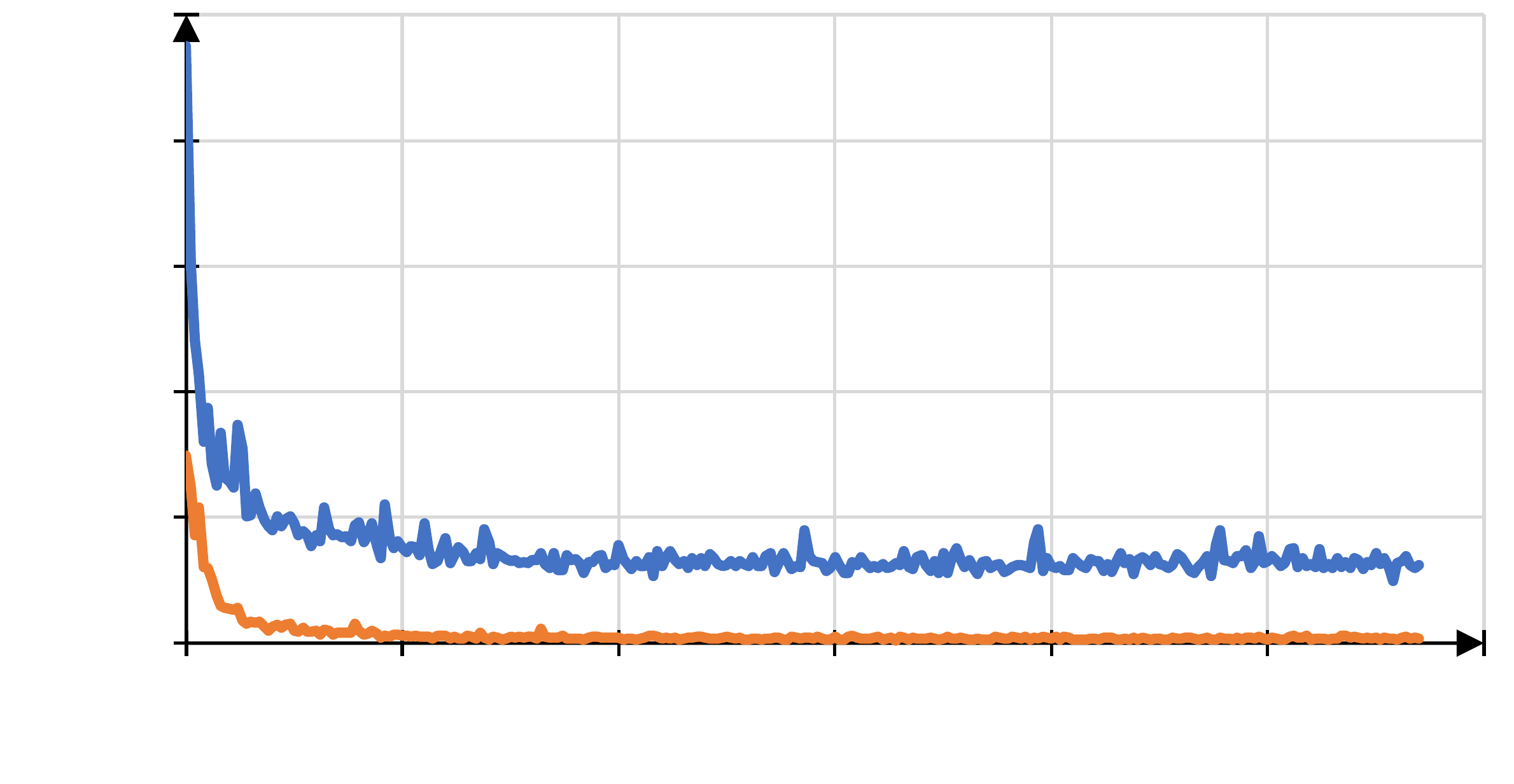}
		\caption{Loss vs Epoch}
		\label{sub:fig:lossepoch}
	\end{subfigure}
	\caption{Training performance}
	\label{fig:training_performance}
\end{figure}

Our neural network implementation utilizes the PyTorch framework \cite{paszke2019pytorch}. We ensure reproducibility by setting a random seed of 42 and using modules such as NumPy and PyTorch. For optimization, we employ the Adam optimizer with a learning rate of $2e-4$, betas set at $(0.9, 0.999)$, an eps value of $1e-8$, and with the amsgrad option disabled. We incorporate a learning rate scheduler and an early stopping strategy to optimize training performance. The "reduce on plateau" learning rate scheduler has a configuration with mode set to minimum, a factor of $0.7$, patience of three epochs, a threshold of $1e-4$, an eps value of $1e-8$, verbose mode disabled, and cooldown and minimum learning rate set to zero. We assign a patience of $40$ epochs and a minimum change in validation loss of 1e-6 for early stopping. If the validation loss doesn't show improvement (change less than $1e-6$) for $40$ consecutive epochs, the training process halts. We save the Model's weights when the validation loss during training is lower than the previous lowest value, thus capturing the best performance achieved during training. These optimization strategies enhance performance and reproducibility in network training, surpassing non-optimized models. This results in more consistent and reliable outcomes. Figure~\ref{fig:training_performance} shows the training performance through these applied optimizations, achieving completion in two days and four hours with 285 epochs.

\subsection{\MakeUppercase{Hardware setup}}
\label{sec:hardwaresetup}
The demonstrator is a light-proof box with an inner dimension of $1200$ x $800$ x $2033$ mm$^3$. It incorporates a clearance height of 402.00 mm at the bottom for an AGV or movable table to pass through. The camera mounting is $1088$ mm above the clearance, ensuring optimal image capturing conditions. A diffuse artificial light source is positioned behind the camera, providing a constant and uniform light distribution throughout the imaging process. A Raspberry Pi 4 computer with a touch display is on the side of the prototype setup. The model trains on an AMD Ryzen 9 7950x 16-Core Processor with 64 GB RAM and Nvidia GeForce RTX 3070 Lite Hash Rate with 8GB VRAM.

\section{\MakeUppercase{Results \& Discussion}}
In the context of classification analysis, the False Positive Rate (FPR) and True Positive Rate (TPR) are pivotal metrics employed to evaluate the efficacy of a classifier in alignment with actual ground truth labels. In the context of an unsupervised model akin to ours, synthetically generated anomalies influence the assessment of the model's performance. It is noteworthy to consider that incorporating bona fide ground truth data could enhance the model's interpretative capacity.

\begin{figure}[H]
	\centering
	\def\svgwidth{0.56\textwidth}
	\resizebox{0.49\textwidth}{!}{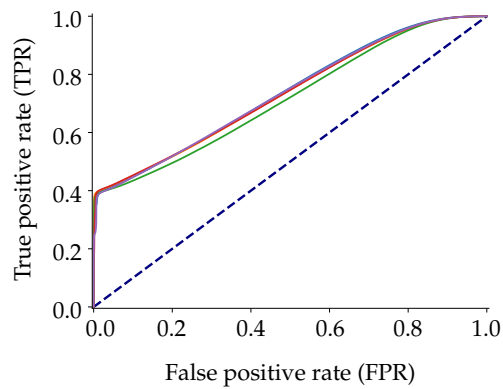}
	\caption{ROC plot of the model predictions on seven different melamine-faced board images, each plotted for a range of threshold values.}
	\label{fig:roc_prototype}
\end{figure}

In our investigation, a suite of seven discrete melamine-faced board images assumes the role of the litmus test for gauging the model's capabilities, portrayed in Figure~\ref{fig:roc_prototype}'s ROC plot. The subsequent narrative endeavors to unfold the nuanced interpretations latent within this ROC plot. The zone where TPR attains unity while FPR dwells at zero demonstrates an impeccable proficiency in demarcating anomalies from their non-anomalous counterparts. However, a discernible pattern emerges in the investigation of the shown ROC plot. Initially, at exceedingly low thresholds, the TPR experiences a precipitous ascent until $0.4$ while the FPR remains close to zero. This dynamic underscores the model's capacity to effectively localize anomalies, albeit with a trade-off that precision remains high while accuracy is fair. Proceeding along this ROC plot, a shift in the equilibrium is palpable. The velocity at which FPR escalates surpasses that of TPR as thresholds ascend. This divergence indicates the model's tendency to misclassify non-anomalous pixels as anomalies. Though accuracy registers an uptick, a proportionate decline in precision is also observed. Such a juxtaposition warrants a meticulous inquiry to study the underlying catalyst — whether inherent noise in the data or a manifestation of model overfitting to specific patterns — contributing to the abrupt spike in FPR. As thresholds scale higher, a unique dynamic manifests. The model traverses the image where all anomalous pixels are successfully localized; however, this feat coexists with a counterpoint — the erroneous labeling of non-anomalous pixels. This paradox underscores a control of heightened accuracy counterbalanced by a degrading precision.

In sum, this investigation of the ROC plot within our context shines a spotlight on the intricate interplay between the True Positive Rate and False Positive Rate, unraveling insights into the model's discrimination prowess, precision, and potential pitfalls attributed to varying thresholds. We select the average threshold value $0.04$, corresponding to the TPR value $0.4$, to obtain accurate predictions while maintaining low misclassification.

As previously stated, the uncropped images are of high resolution, and most of the anomalies are relatively tiny compared to the size of the other features. To better visualize the defects, the evaluation focuses on identifying specific crops of the melamine-faced board to showcase the performance, encompassing corners, edges, grooves, holes, and surfaces with their actual sharpness. Magnifying Figure~\ref{fig:results_all} can provide a better visualization of the anomalies. While correct predictions generally characterize the non-anomalous area, accuracy fluctuates amidst anomalous regions. Given an unsupervised defect localizing model, achieving precise pixel-level performance is challenging due to the lack of definitive ground truth for the model to learn. Addressing this is done by generating heatmaps from the difference between the original and reconstructed images. These heatmaps are then overlaid with the corresponding actual anomalous crop areas to gauge the model behavior and the quality of its localization (refer to Figure~\ref{fig:results_all}). Each tile features a $289 \text{ x } 289$ resolution crop, organized in sets of three (\textit{a}, \textit{b}, and \textit{c}) in each of the four columns. The first tile, \textit{a}, represents the actual anomalous area crop, followed by the second tile, \textit{b}, displaying the heatmap resulting from the difference, and the third tile, \textit{c}, presenting an overlay of the heatmap on the actual anomalous crop area, with opacities set at 75\% and 50\% respectively.

The model demonstrates decent capabilities regarding localization of tiny artifacts such as smudges, dirt, and deformities on the corners of the melamine-faced boards in Figure~\hyperref[fig:sub:p_b_2_3]{\ref{fig:results_all}A1}, Figure~\hyperref[fig:sub:p_b_3_11]{\ref{fig:results_all}A3}, and Figure~\hyperref[fig:sub:p_b_3_12]{\ref{fig:results_all}A4}. The model accurately identifies these imperfections. However, the model’s performance falls short when predicting the presence of large imperfections on the melamine-faced boards, as evidenced in Figure~\hyperref[fig:sub:p_b_2_7]{\ref{fig:results_all}A2}. It appears that the model struggles to detect and localize large-scale defects.

In the context of anomaly localization close to the edges of the boards, as shown in Figure~\hyperref[fig:sub:B]{\ref{fig:results_all}B - C}, the model’s performance in predicting significant defects of various shapes and sizes is decent. However, challenges arise when dealing with relatively small-scale or (and) blurred anomalies, such as the one located on the left edge in Figure~\hyperref[fig:sub:p_b_1_3]{\ref{fig:results_all}B2}. In such cases, the model’s confidence in its prediction decreases, leading to less accurate results. Figure~\hyperref[fig:sub:p_b_4_3]{\ref{fig:results_all}C1} showcases the model’s capability to accurately localize artifacts even when positioned slightly away from the edge, demonstrating its potential for robust anomaly detection in various scenarios.

The model exhibits mixed confidence when dealing with artifacts around the groove, as evidenced in Figure~\hyperref[fig:sub:D]{\ref{fig:results_all}D1 - D4}. This behavior can be attributed to the artifact closely resembling the surrounding environment, making it challenging for the model to distinguish it as an anomaly. However, the model’s performance receives a significant boost when dealing with instances of discontinuity. Discontinuities in the data are more straightforward for the model to identify and classify as anomalies, resulting in higher confidence predictions. Moreover, the defects found at the end of the groove are also localized by the model with moderate confidence, as seen in Figure~\hyperref[fig:sub:p_b_6_24]{\ref{fig:results_all}E1}, suggesting that the model can detect these defects to some extent, but there may still be room for improvement in accuracy and precision.

The model’s performance in localizing anomalies around holes is observable in Figure~\hyperref[fig:sub:F]{\ref{fig:results_all}F - H}. It successfully identifies anomalies within hole textures. Moreover, the model exhibits its proficiency in accurately pinpointing localizing different types of defects, such as defects along edges and holes (Figure~\hyperref[fig:sub:p_b_1_7]{\ref{fig:results_all}F4}) as well as defects on surfaces and holes (evident in Figure~\hyperref[fig:sub:p_b_1_8]{\ref{fig:results_all}G1}, Figure~\hyperref[fig:sub:p_b_1_10]{\ref{fig:results_all}G3}, and Figure~\hyperref[fig:sub:p_b_5_22]{\ref{fig:results_all}H1}). The versatile capability to address diverse anomaly types underscores the model’s effectiveness and potential.

Figure~\hyperref[fig:sub:I]{\ref{fig:results_all}I - N} evidences the model’s localization performance for plain surface defects. It successfully detects prominent anomalies and localizes even the most minute plain surface defects, as evident in Figure~\hyperref[fig:sub:p_b_0_7]{\ref{fig:results_all}J1 - J4}. Moreover, the model demonstrates its ability to recognize defects resembling closely to holes in shape and texture, as observed in Figure~\hyperref[fig:sub:p_b_5_10]{\ref{fig:results_all}M1}.

Overall, the results in Figure~\ref{fig:results_all} highlight the model's robustness, limitations, and accuracy in tackling complex defect localization tasks by identifying and localizing a wide range of surface defects occurring on melamine-faced boards.

\newcommand{\reshspace}{0.4em}
\newcommand{\indexwidth}{0.02\textwidth}
\newcommand{\inclwidth}{0.98\textwidth}
\newcommand{\hshead}{1.7em}
\begin{figure}
	\centering
	\begin{subfigure}{\textwidth}
		\begin{minipage}{\indexwidth}
			\centering
			
		\end{minipage}
		\hspace{0.5cm}
		\begin{minipage}{\indexwidth}
			\centering
			1a
		\end{minipage}
		\hspace{\hshead}
		\begin{minipage}{\indexwidth}
			\centering
			1b  
		\end{minipage}
		\hspace{\hshead}
		\begin{minipage}{\indexwidth}
			\centering
			1c  
		\end{minipage}
		\hspace{0.6em}
		\hspace{\hshead}
		\begin{minipage}{\indexwidth}
			\centering
			2a
		\end{minipage}
		\hspace{\hshead}
		\begin{minipage}{\indexwidth}
			\centering
			2b  
		\end{minipage}
		\hspace{\hshead}
		\begin{minipage}{\indexwidth}
			\centering
			2c  
		\end{minipage}
		\hspace{0.6em}
		\hspace{\hshead}
		\begin{minipage}{\indexwidth}
			\centering
			3a
		\end{minipage}
		\hspace{\hshead}
		\begin{minipage}{\indexwidth}
			\centering
			3b  
		\end{minipage}
		\hspace{\hshead}
		\begin{minipage}{\indexwidth}
			\centering
			3c  
		\end{minipage}
		\hspace{0.6em}
		\hspace{\hshead}
		\begin{minipage}{\indexwidth}
			\centering
			4a
		\end{minipage}
		\hspace{\hshead}
		\begin{minipage}{\indexwidth}
			\centering
			4b  
		\end{minipage}
		\hspace{\hshead}
		\begin{minipage}{\indexwidth}
			\centering
			4c  
		\end{minipage}
	\end{subfigure}
	\begin{subfigure}{\textwidth}
		\begin{minipage}{\indexwidth}
			\centering
			A
			\label{fig:sub:A}
		\end{minipage}
		\begin{minipage}{\inclwidth}
			\begin{subfigure}{0.0725\textwidth}
				\centering
				\includegraphics[width=\linewidth]{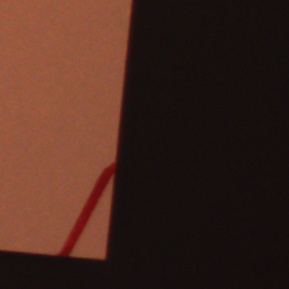}
				\label{fig:sub:p_b_2_3}
			\end{subfigure}
			\begin{subfigure}{0.0725\textwidth}
				\centering
				\includegraphics[width=\linewidth]{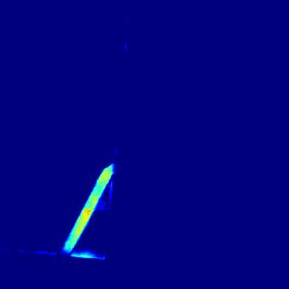}
				\label{fig:sub:p_h_2_3}
			\end{subfigure}
			\begin{subfigure}{0.0725\textwidth}
				\centering
				\includegraphics[width=\linewidth]{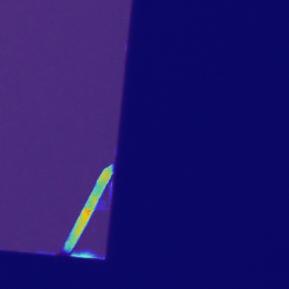}
				\label{fig:sub:p_o_2_3}
			\end{subfigure}
			\hspace{\reshspace}
			\begin{subfigure}{0.0725\textwidth}
				\centering
				\includegraphics[width=\linewidth]{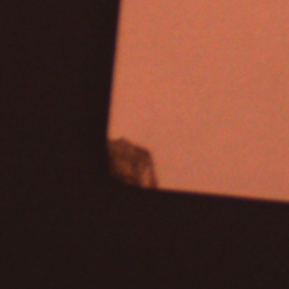}
				\label{fig:sub:p_b_2_7}
			\end{subfigure}
			\begin{subfigure}{0.0725\textwidth}
				\centering
				\includegraphics[width=\linewidth]{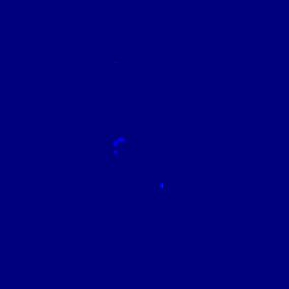}
				\label{fig:sub:p_h_2_7}
			\end{subfigure}
			\begin{subfigure}{0.0725\textwidth}
				\centering
				\includegraphics[width=\linewidth]{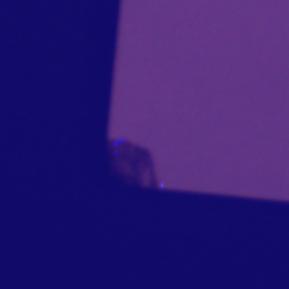}
				\label{fig:sub:p_o_2_7}
			\end{subfigure}
			\hspace{\reshspace}
			\begin{subfigure}{0.0725\textwidth}
				\centering
				\includegraphics[width=\linewidth]{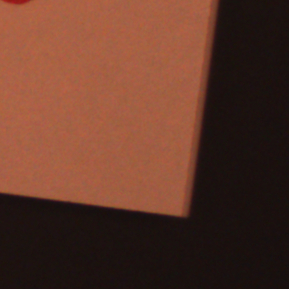}
				\label{fig:sub:p_b_3_11}
			\end{subfigure}
			\begin{subfigure}{0.0725\textwidth}
				\centering
				\includegraphics[width=\linewidth]{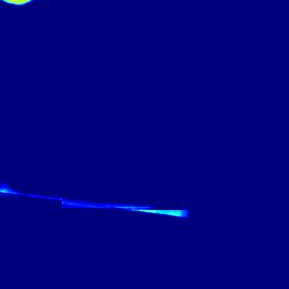}
				\label{fig:sub:p_h_3_11}
			\end{subfigure}
			\begin{subfigure}{0.0725\textwidth}
				\centering
				\includegraphics[width=\linewidth]{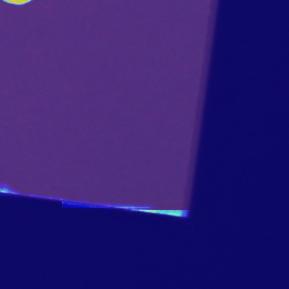}
				\label{fig:sub:p_o_3_11}
			\end{subfigure}
			\hspace{\reshspace}
			\begin{subfigure}{0.0725\textwidth}
				\centering
				\includegraphics[width=\linewidth]{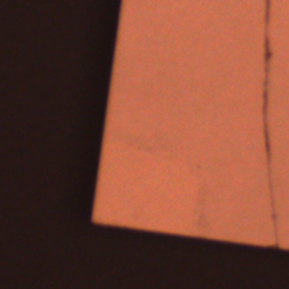}
				\label{fig:sub:p_b_3_12}
			\end{subfigure}
			\begin{subfigure}{0.0725\textwidth}
				\centering
				\includegraphics[width=\linewidth]{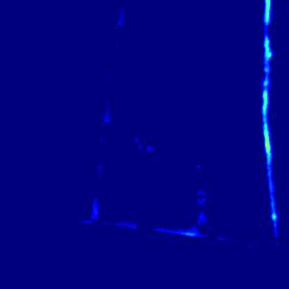}
				\label{fig:sub:p_h_3_12}
			\end{subfigure}
			\begin{subfigure}{0.0725\textwidth}
				\centering
				\includegraphics[width=\linewidth]{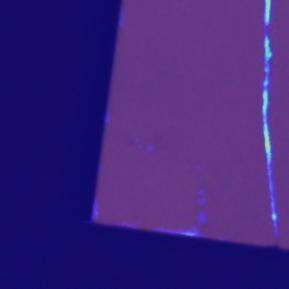}
				\label{fig:sub:p_o_3_12}
			\end{subfigure}
			\vspace{-0.35cm}
		\end{minipage}
	\end{subfigure}
	\begin{subfigure}{\textwidth}
		\begin{minipage}{\indexwidth}
			\centering
			B
			\label{fig:sub:B}
		\end{minipage}
		\begin{minipage}{\inclwidth}
			\begin{subfigure}{0.0725\textwidth}
				\centering
				\includegraphics[width=\linewidth]{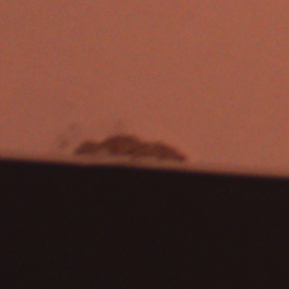}
				\label{fig:sub:p_b_0_16}
			\end{subfigure}
			\begin{subfigure}{0.0725\textwidth}
				\centering
				\includegraphics[width=\linewidth]{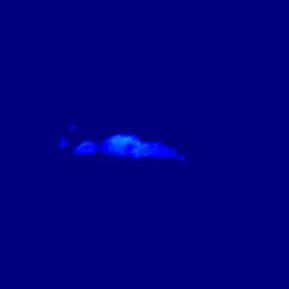}
				\label{fig:sub:p_h_0_16}
			\end{subfigure}
			\begin{subfigure}{0.0725\textwidth}
				\centering
				\includegraphics[width=\linewidth]{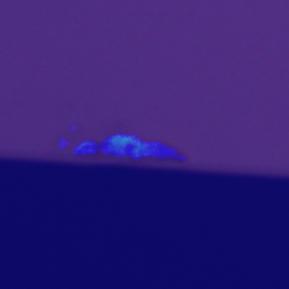}
				\label{fig:sub:p_o_0_16}
			\end{subfigure}
			\hspace{\reshspace}
			\begin{subfigure}{0.0725\textwidth}
				\centering
				\includegraphics[width=\linewidth]{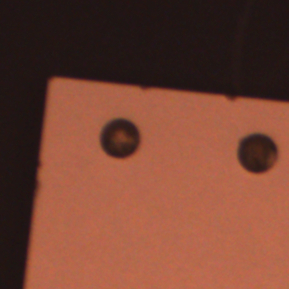}
				\label{fig:sub:p_b_1_3}
			\end{subfigure}
			\begin{subfigure}{0.0725\textwidth}
				\centering
				\includegraphics[width=\linewidth]{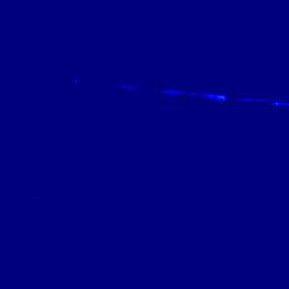}
				\label{fig:sub:p_h_1_3}
			\end{subfigure}
			\begin{subfigure}{0.0725\textwidth}
				\centering
				\includegraphics[width=\linewidth]{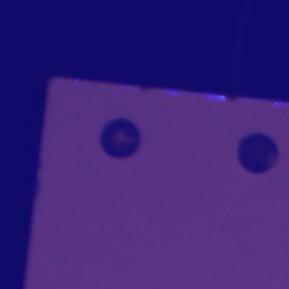}
				\label{fig:sub:p_o_1_3}
			\end{subfigure}
			\hspace{\reshspace}
			\begin{subfigure}{0.0725\textwidth}
				\centering
				\includegraphics[width=\linewidth]{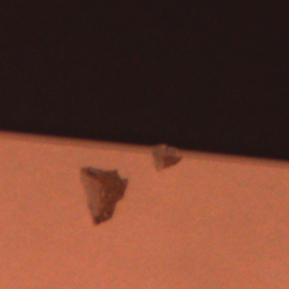}
				\label{fig:sub:p_b_2_4}
			\end{subfigure}
			\begin{subfigure}{0.0725\textwidth}
				\centering
				\includegraphics[width=\linewidth]{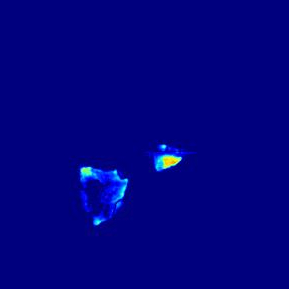}
				\label{fig:sub:p_h_2_4}
			\end{subfigure}
			\begin{subfigure}{0.0725\textwidth}
				\centering
				\includegraphics[width=\linewidth]{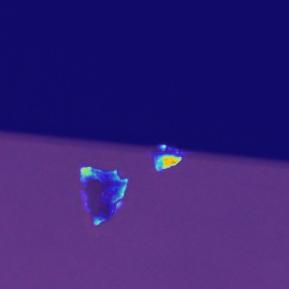}
				\label{fig:sub:p_o_2_4}
			\end{subfigure}
			\hspace{\reshspace}
			\begin{subfigure}{0.0725\textwidth}
				\centering
				\includegraphics[width=\linewidth]{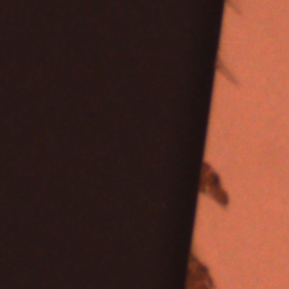}
				\label{fig:sub:p_b_3_9}
			\end{subfigure}
			\begin{subfigure}{0.0725\textwidth}
				\centering
				\includegraphics[width=\linewidth]{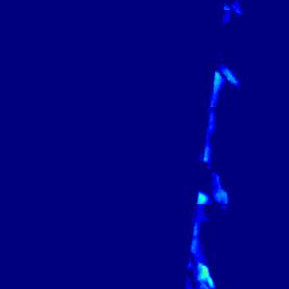}
				\label{fig:sub:p_h_3_9}
			\end{subfigure}
			\begin{subfigure}{0.0725\textwidth}
				\centering
				\includegraphics[width=\linewidth]{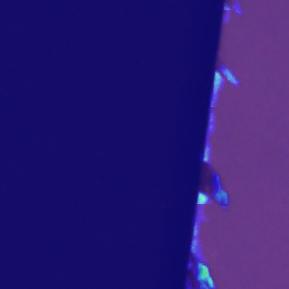}
				\label{fig:sub:p_o_3_9}
			\end{subfigure}
			\vspace{-0.35cm}
		\end{minipage}
	\end{subfigure}
	\begin{subfigure}{\textwidth}
		\begin{minipage}{\indexwidth}
			\centering
			C
			\label{fig:sub:C}
		\end{minipage}
		\begin{minipage}{\inclwidth}
			\begin{subfigure}{0.0725\textwidth}
				\centering
				\includegraphics[width=\linewidth]{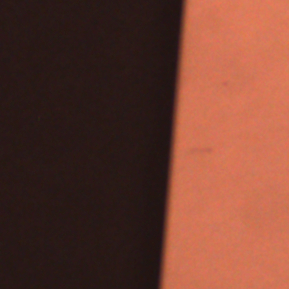}
				\label{fig:sub:p_b_4_3}
			\end{subfigure}
			\begin{subfigure}{0.0725\textwidth}
				\centering
				\includegraphics[width=\linewidth]{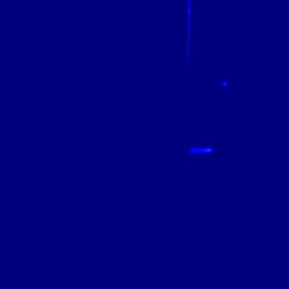}
				\label{fig:sub:p_h_4_3}
			\end{subfigure}
			\begin{subfigure}{0.0725\textwidth}
				\centering
				\includegraphics[width=\linewidth]{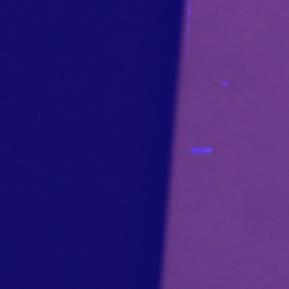}
				\label{fig:sub:p_o_4_3}
			\end{subfigure}
			\vspace{-0.35cm}
		\end{minipage}
	\end{subfigure}
	\begin{subfigure}{\textwidth}
		\begin{minipage}{\indexwidth}
			\centering
			D
			\label{fig:sub:D}
		\end{minipage}
		\begin{minipage}{\inclwidth}
			\begin{subfigure}{0.0725\textwidth}
				\centering
				\includegraphics[width=\linewidth]{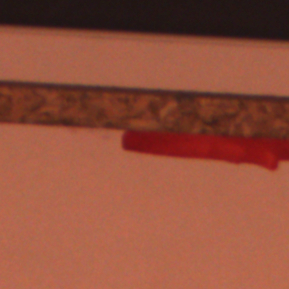}
				\label{fig:sub:p_b_0_17}
			\end{subfigure}
			\begin{subfigure}{0.0725\textwidth}
				\centering
				\includegraphics[width=\linewidth]{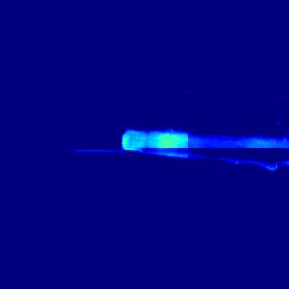}
				\label{fig:sub:p_h_0_17}
			\end{subfigure}
			\begin{subfigure}{0.0725\textwidth}
				\centering
				\includegraphics[width=\linewidth]{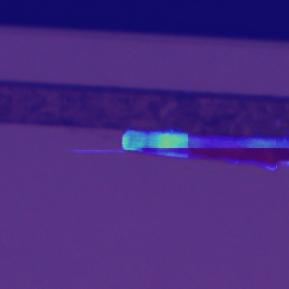}
				\label{fig:sub:p_o_0_17}
			\end{subfigure}
			\hspace{\reshspace}
			\begin{subfigure}{0.0725\textwidth}
				\centering
				\includegraphics[width=\linewidth]{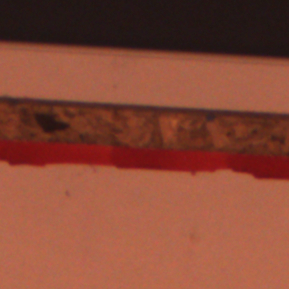}
				\label{fig:sub:p_b_0_18}
			\end{subfigure}
			\begin{subfigure}{0.0725\textwidth}
				\centering
				\includegraphics[width=\linewidth]{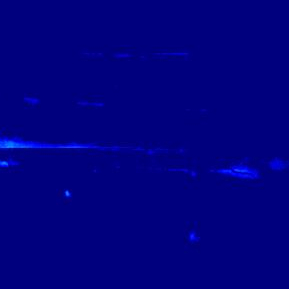}
				\label{fig:sub:p_h_0_18}
			\end{subfigure}
			\begin{subfigure}{0.0725\textwidth}
				\centering
				\includegraphics[width=\linewidth]{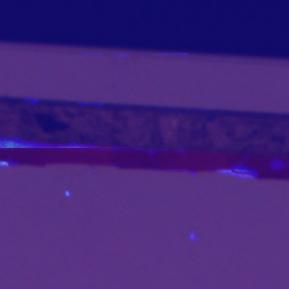}
				\label{fig:sub:p_o_0_18}
			\end{subfigure}
			\hspace{\reshspace}
			\begin{subfigure}{0.0725\textwidth}
				\centering
				\includegraphics[width=\linewidth]{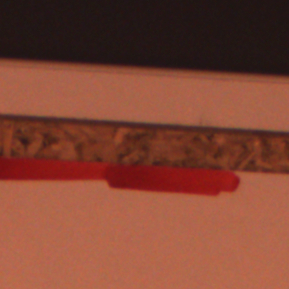}
				\label{fig:sub:p_b_0_19}
			\end{subfigure}
			\begin{subfigure}{0.0725\textwidth}
				\centering
				\includegraphics[width=\linewidth]{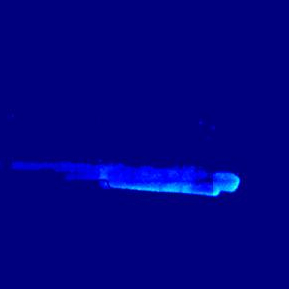}
				\label{fig:sub:p_h_0_19}
			\end{subfigure}
			\begin{subfigure}{0.0725\textwidth}
				\centering
				\includegraphics[width=\linewidth]{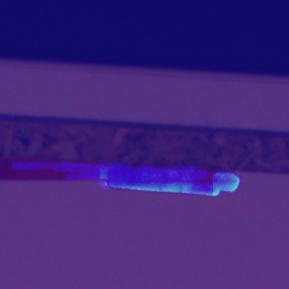}
				\label{fig:sub:p_o_0_19}
			\end{subfigure}
			\hspace{\reshspace}
			\begin{subfigure}{0.0725\textwidth}
				\centering
				\includegraphics[width=\linewidth]{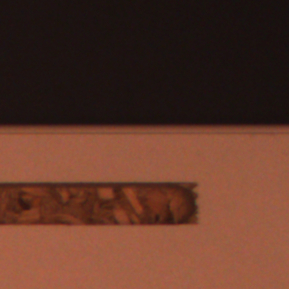}
				\label{fig:sub:p_b_5_3}
			\end{subfigure}
			\begin{subfigure}{0.0725\textwidth}
				\centering
				\includegraphics[width=\linewidth]{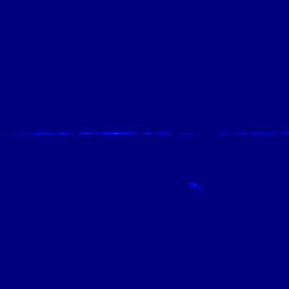}
				\label{fig:sub:p_h_5_3}
			\end{subfigure}
			\begin{subfigure}{0.0725\textwidth}
				\centering
				\includegraphics[width=\linewidth]{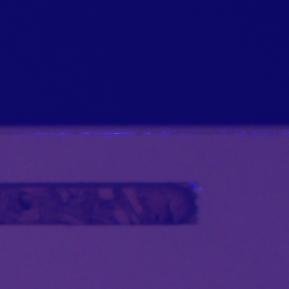}
				\label{fig:sub:p_o_5_3}
			\end{subfigure}
			\vspace{-0.35cm}
		\end{minipage}
	\end{subfigure}
	\begin{subfigure}{\textwidth}
		\begin{minipage}{\indexwidth}
			\centering
			E
			\label{fig:sub:E}
		\end{minipage}
		\begin{minipage}{\inclwidth}
			\begin{subfigure}{0.0725\textwidth}
				\centering
				\includegraphics[width=\linewidth]{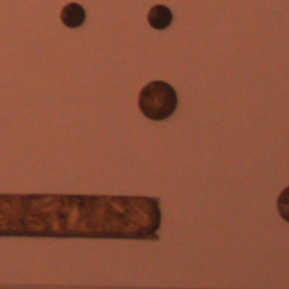}
				\label{fig:sub:p_b_6_24}
			\end{subfigure}
			\begin{subfigure}{0.0725\textwidth}
				\centering
				\includegraphics[width=\linewidth]{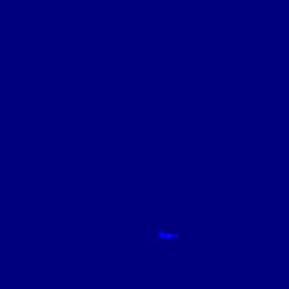}
				\label{fig:sub:p_h_6_24}
			\end{subfigure}
			\begin{subfigure}{0.0725\textwidth}
				\centering
				\includegraphics[width=\linewidth]{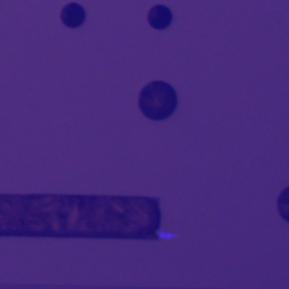}
				\label{fig:sub:p_o_6_24}
			\end{subfigure}
			\vspace{-0.35cm}
		\end{minipage}
	\end{subfigure}
	\begin{subfigure}{\textwidth}
		\begin{minipage}{\indexwidth}
			\centering
			F
			\label{fig:sub:F}
		\end{minipage}
		\begin{minipage}{\inclwidth}
			\begin{subfigure}{0.0725\textwidth}
				\centering
				\includegraphics[width=\linewidth]{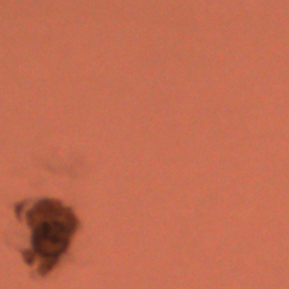}
				\label{fig:sub:p_b_0_5}
			\end{subfigure}
			\begin{subfigure}{0.0725\textwidth}
				\centering
				\includegraphics[width=\linewidth]{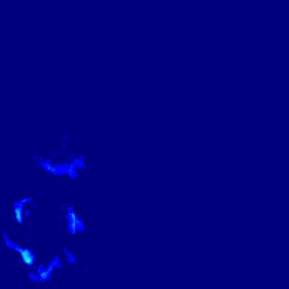}
				\label{fig:sub:p_h_0_5}
			\end{subfigure}
			\begin{subfigure}{0.0725\textwidth}
				\centering
				\includegraphics[width=\linewidth]{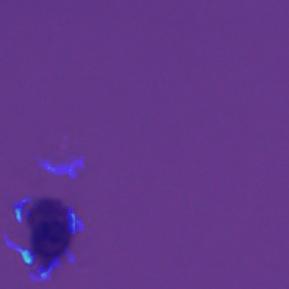}
				\label{fig:sub:p_o_0_5}
			\end{subfigure}
			\hspace{\reshspace}
			\begin{subfigure}{0.0725\textwidth}
				\centering
				\includegraphics[width=\linewidth]{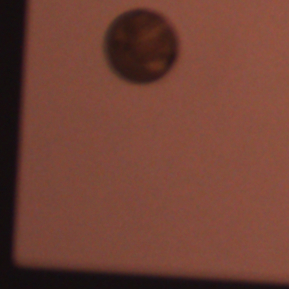}
				\label{fig:sub:p_b_0_10}
			\end{subfigure}
			\begin{subfigure}{0.0725\textwidth}
				\centering
				\includegraphics[width=\linewidth]{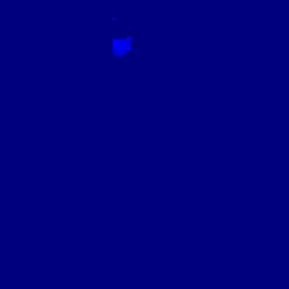}
				\label{fig:sub:p_h_0_10}
			\end{subfigure}
			\begin{subfigure}{0.0725\textwidth}
				\centering
				\includegraphics[width=\linewidth]{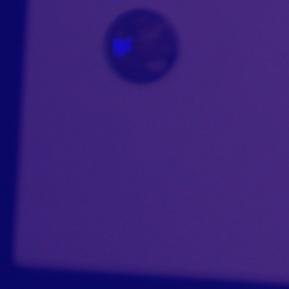}
				\label{fig:sub:p_o_0_10}
			\end{subfigure}
			\hspace{\reshspace}
			\begin{subfigure}{0.0725\textwidth}
				\centering
				\includegraphics[width=\linewidth]{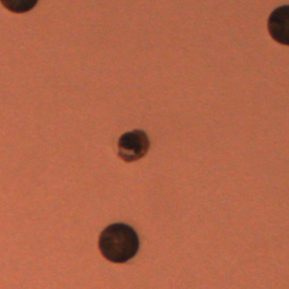}
				\label{fig:sub:p_b_1_1}
			\end{subfigure}
			\begin{subfigure}{0.0725\textwidth}
				\centering
				\includegraphics[width=\linewidth]{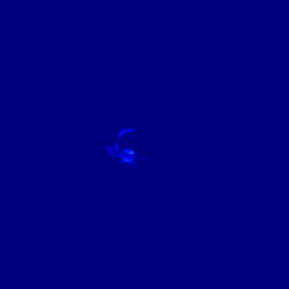}
				\label{fig:sub:p_h_1_1}
			\end{subfigure}
			\begin{subfigure}{0.0725\textwidth}
				\centering
				\includegraphics[width=\linewidth]{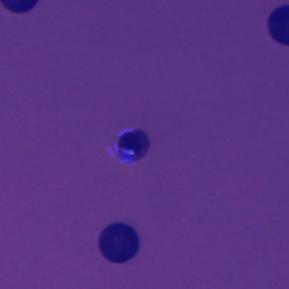}
				\label{fig:sub:p_o_1_1}
			\end{subfigure}
			\hspace{\reshspace}
			\begin{subfigure}{0.0725\textwidth}
				\centering
				\includegraphics[width=\linewidth]{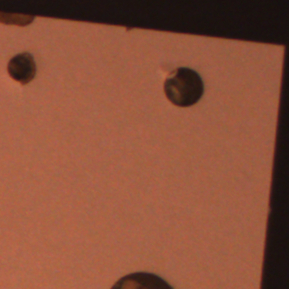}
				\label{fig:sub:p_b_1_7}
			\end{subfigure}
			\begin{subfigure}{0.0725\textwidth}
				\centering
				\includegraphics[width=\linewidth]{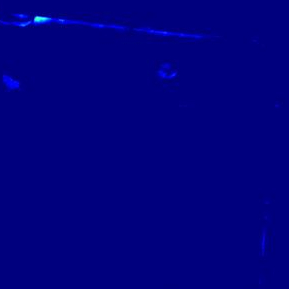}
				\label{fig:sub:p_h_1_7}
			\end{subfigure}
			\begin{subfigure}{0.0725\textwidth}
				\centering
				\includegraphics[width=\linewidth]{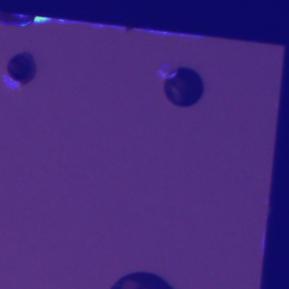}
				\label{fig:sub:p_o_1_7}
			\end{subfigure}
			\vspace{-0.35cm}
		\end{minipage}
	\end{subfigure}
	\begin{subfigure}{\textwidth}
		\begin{minipage}{\indexwidth}
			\centering
			G
			\label{fig:sub:G}
		\end{minipage}
		\begin{minipage}{\inclwidth}
			\begin{subfigure}{0.0725\textwidth}
				\centering
				\includegraphics[width=\linewidth]{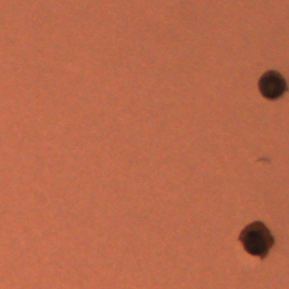}
				\label{fig:sub:p_b_1_8}
			\end{subfigure}
			\begin{subfigure}{0.0725\textwidth}
				\centering
				\includegraphics[width=\linewidth]{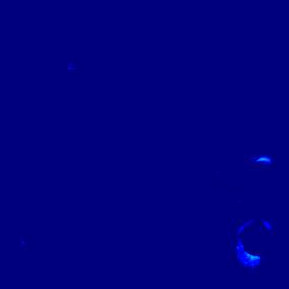}
				\label{fig:sub:p_h_1_8}
			\end{subfigure}
			\begin{subfigure}{0.0725\textwidth}
				\centering
				\includegraphics[width=\linewidth]{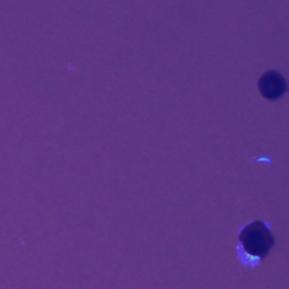}
				\label{fig:sub:p_o_1_8}
			\end{subfigure}
			\hspace{\reshspace}
			\begin{subfigure}{0.0725\textwidth}
				\centering
				\includegraphics[width=\linewidth]{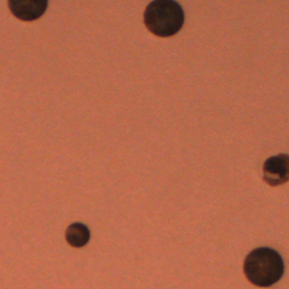}
				\label{fig:sub:p_b_1_9}
			\end{subfigure}
			\begin{subfigure}{0.0725\textwidth}
				\centering
				\includegraphics[width=\linewidth]{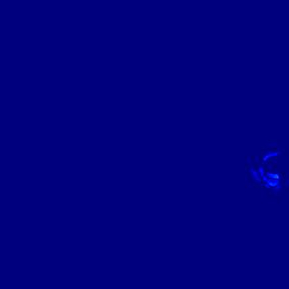}
				\label{fig:sub:p_h_1_9}
			\end{subfigure}
			\begin{subfigure}{0.0725\textwidth}
				\centering
				\includegraphics[width=\linewidth]{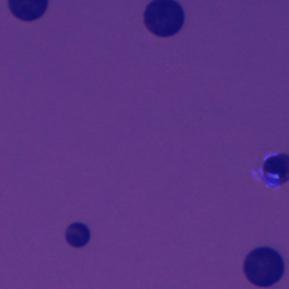}
				\label{fig:sub:p_o_1_9}
			\end{subfigure}
			\hspace{\reshspace}
			\begin{subfigure}{0.0725\textwidth}
				\centering
				\includegraphics[width=\linewidth]{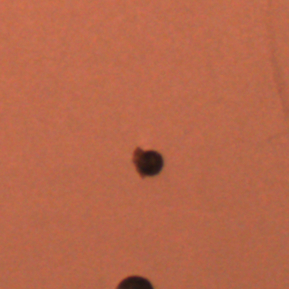}
				\label{fig:sub:p_b_1_10}
			\end{subfigure}
			\begin{subfigure}{0.0725\textwidth}
				\centering
				\includegraphics[width=\linewidth]{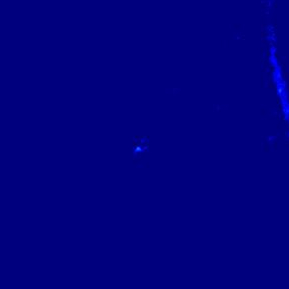}
				\label{fig:sub:p_h_1_10}
			\end{subfigure}
			\begin{subfigure}{0.0725\textwidth}
				\centering
				\includegraphics[width=\linewidth]{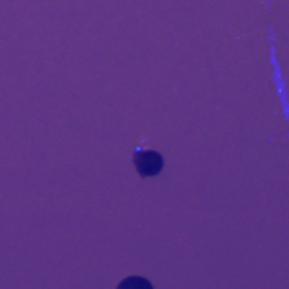}
				\label{fig:sub:p_o_1_10}
			\end{subfigure}
			\hspace{\reshspace}
			\begin{subfigure}{0.0725\textwidth}
				\centering
				\includegraphics[width=\linewidth]{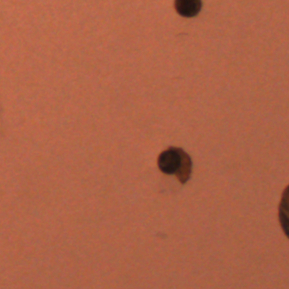}
				\label{fig:sub:p_b_1_11}
			\end{subfigure}
			\begin{subfigure}{0.0725\textwidth}
				\centering
				\includegraphics[width=\linewidth]{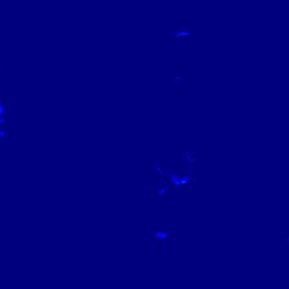}
				\label{fig:sub:p_h_1_11}
			\end{subfigure}
			\begin{subfigure}{0.0725\textwidth}
				\centering
				\includegraphics[width=\linewidth]{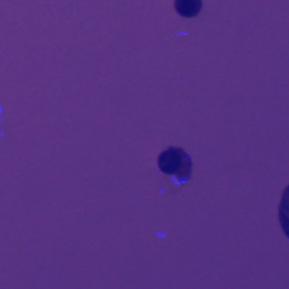}
				\label{fig:sub:p_o_1_11}
			\end{subfigure}
			\vspace{-0.35cm}
		\end{minipage}
	\end{subfigure}
	\begin{subfigure}{\textwidth}
		\begin{minipage}{\indexwidth}
			\centering
			H
			\label{fig:sub:H}
		\end{minipage}
		\begin{minipage}{\inclwidth}
			\begin{subfigure}{0.0725\textwidth}
				\centering
				\includegraphics[width=\linewidth]{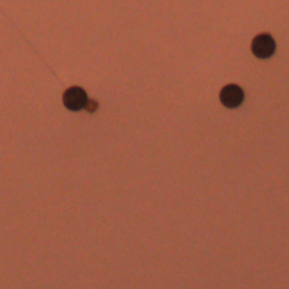}
				\label{fig:sub:p_b_5_22}
			\end{subfigure}
			\begin{subfigure}{0.0725\textwidth}
				\centering
				\includegraphics[width=\linewidth]{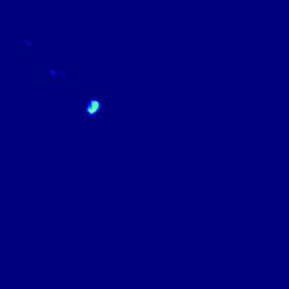}
				\label{fig:sub:p_h_5_22}
			\end{subfigure}
			\begin{subfigure}{0.0725\textwidth}
				\centering
				\includegraphics[width=\linewidth]{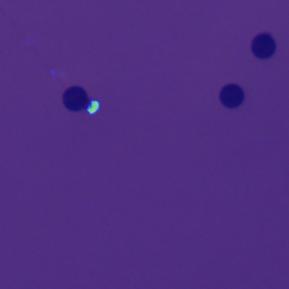}
				\label{fig:sub:p_o_5_22}
			\end{subfigure}
			\hspace{\reshspace}
			\begin{subfigure}{0.0725\textwidth}
				\centering
				\includegraphics[width=\linewidth]{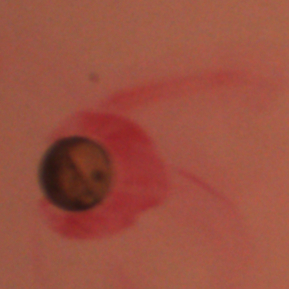}
				\label{fig:sub:p_b_6_5}
			\end{subfigure}
			\begin{subfigure}{0.0725\textwidth}
				\centering
				\includegraphics[width=\linewidth]{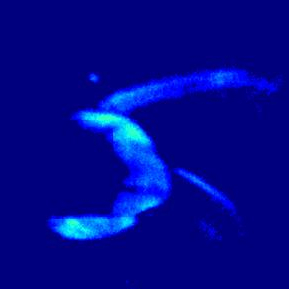}
				\label{fig:sub:p_h_6_5}
			\end{subfigure}
			\begin{subfigure}{0.0725\textwidth}
				\centering
				\includegraphics[width=\linewidth]{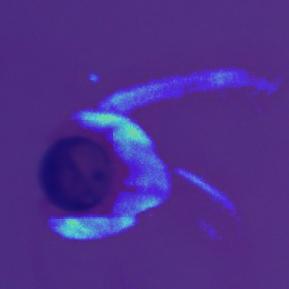}
				\label{fig:sub:p_o_6_5}
			\end{subfigure}
			\hspace{\reshspace}
			\begin{subfigure}{0.0725\textwidth}
				\centering
				\includegraphics[width=\linewidth]{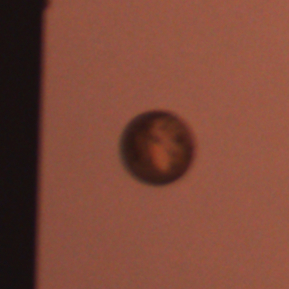}
				\label{fig:sub:p_b_6_15}
			\end{subfigure}
			\begin{subfigure}{0.0725\textwidth}
				\centering
				\includegraphics[width=\linewidth]{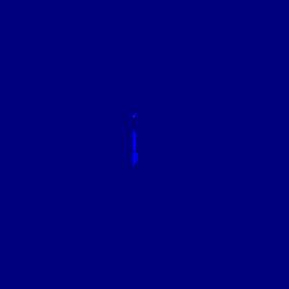}
				\label{fig:sub:p_h_6_15}
			\end{subfigure}
			\begin{subfigure}{0.0725\textwidth}
				\centering
				\includegraphics[width=\linewidth]{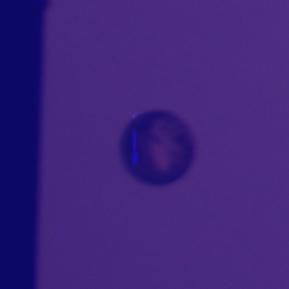}
				\label{fig:sub:p_o_6_15}
			\end{subfigure}
			\vspace{-0.35cm}
		\end{minipage}
	\end{subfigure}
	\begin{subfigure}{\textwidth}
		\begin{minipage}{\indexwidth}
			\centering
			I
			\label{fig:sub:I}
		\end{minipage}
		\begin{minipage}{\inclwidth}
			\begin{subfigure}{0.0725\textwidth}
				\centering
				\includegraphics[width=\linewidth]{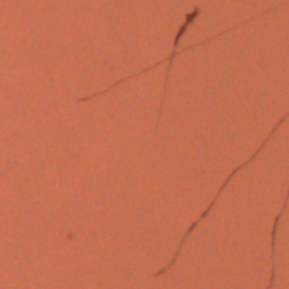}
				\label{fig:sub:p_b_0_1}
			\end{subfigure}
			\begin{subfigure}{0.0725\textwidth}
				\centering
				\includegraphics[width=\linewidth]{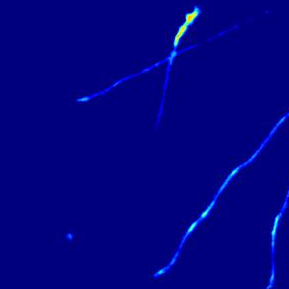}
				\label{fig:sub:p_h_0_1}
			\end{subfigure}
			\begin{subfigure}{0.0725\textwidth}
				\centering
				\includegraphics[width=\linewidth]{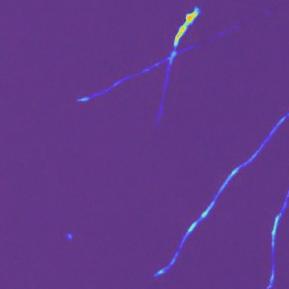}
				\label{fig:sub:p_o_0_1}
			\end{subfigure}
			\hspace{\reshspace}
			\begin{subfigure}{0.0725\textwidth}
				\centering
				\includegraphics[width=\linewidth]{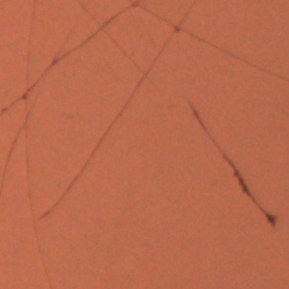}
				\label{fig:sub:p_b_0_2}
			\end{subfigure}
			\begin{subfigure}{0.0725\textwidth}
				\centering
				\includegraphics[width=\linewidth]{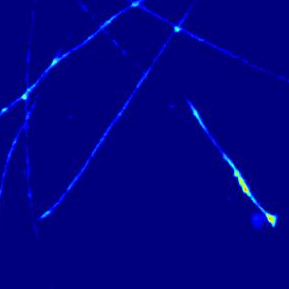}
				\label{fig:sub:p_h_0_2}
			\end{subfigure}
			\begin{subfigure}{0.0725\textwidth}
				\centering
				\includegraphics[width=\linewidth]{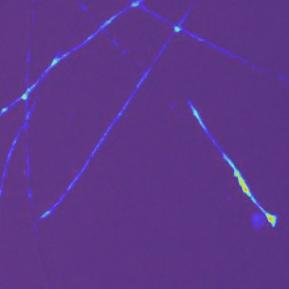}
				\label{fig:sub:p_o_0_2}
			\end{subfigure}
			\hspace{\reshspace}
			\begin{subfigure}{0.0725\textwidth}
				\centering
				\includegraphics[width=\linewidth]{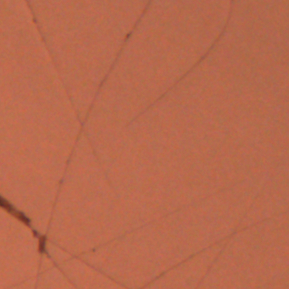}
				\label{fig:sub:p_b_0_3}
			\end{subfigure}
			\begin{subfigure}{0.0725\textwidth}
				\centering
				\includegraphics[width=\linewidth]{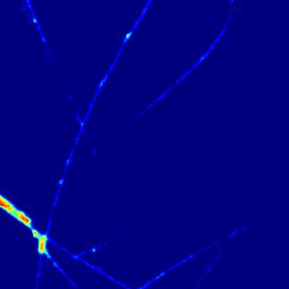}
				\label{fig:sub:p_h_0_3}
			\end{subfigure}
			\begin{subfigure}{0.0725\textwidth}
				\centering
				\includegraphics[width=\linewidth]{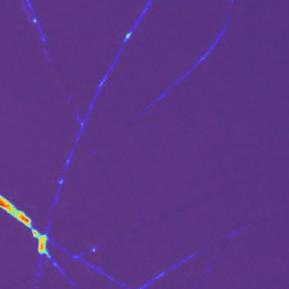}
				\label{fig:sub:p_o_0_3}
			\end{subfigure}
			\hspace{\reshspace}
			\begin{subfigure}{0.0725\textwidth}
				\centering
				\includegraphics[width=\linewidth]{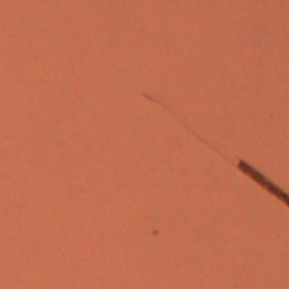}
				\label{fig:sub:p_b_0_4}
			\end{subfigure}
			\begin{subfigure}{0.0725\textwidth}
				\centering
				\includegraphics[width=\linewidth]{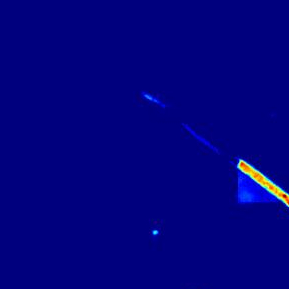}
				\label{fig:sub:p_h_0_4}
			\end{subfigure}
			\begin{subfigure}{0.0725\textwidth}
				\centering
				\includegraphics[width=\linewidth]{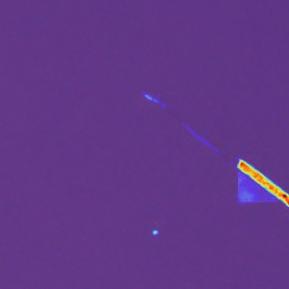}
				\label{fig:sub:p_o_0_4}
			\end{subfigure}
			\vspace{-0.35cm}
		\end{minipage}
	\end{subfigure}
	\begin{subfigure}{\textwidth}
		\begin{minipage}{\indexwidth}
			\centering
			J
			\label{fig:sub:J}
		\end{minipage}
		\begin{minipage}{\inclwidth}
			\begin{subfigure}{0.0725\textwidth}
				\centering
				\includegraphics[width=\linewidth]{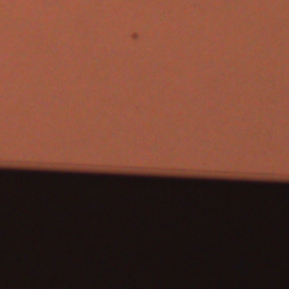}
				\label{fig:sub:p_b_0_7}
			\end{subfigure}
			\begin{subfigure}{0.0725\textwidth}
				\centering
				\includegraphics[width=\linewidth]{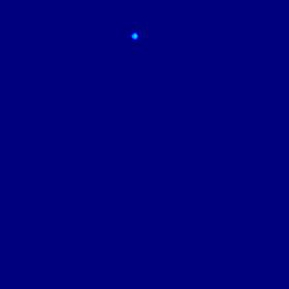}
				\label{fig:sub:p_h_0_7}
			\end{subfigure}
			\begin{subfigure}{0.0725\textwidth}
				\centering
				\includegraphics[width=\linewidth]{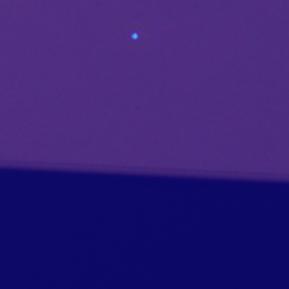}
				\label{fig:sub:p_o_0_7}
			\end{subfigure}
			\hspace{\reshspace}
			\begin{subfigure}{0.0725\textwidth}
				\centering
				\includegraphics[width=\linewidth]{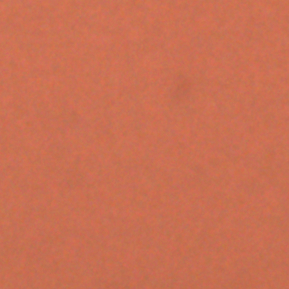}
				\label{fig:sub:p_b_1_16}
			\end{subfigure}
			\begin{subfigure}{0.0725\textwidth}
				\centering
				\includegraphics[width=\linewidth]{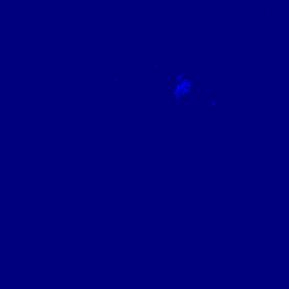}
				\label{fig:sub:p_h_1_16}
			\end{subfigure}
			\begin{subfigure}{0.0725\textwidth}
				\centering
				\includegraphics[width=\linewidth]{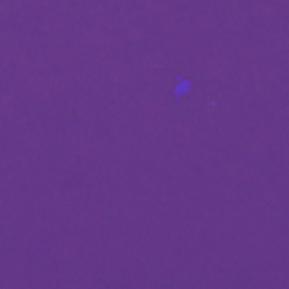}
				\label{fig:sub:p_o_1_16}
			\end{subfigure}
			\hspace{\reshspace}
			\begin{subfigure}{0.0725\textwidth}
				\centering
				\includegraphics[width=\linewidth]{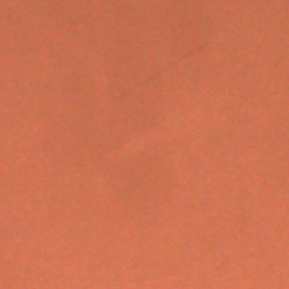}
				\label{fig:sub:p_b_2_5}
			\end{subfigure}
			\begin{subfigure}{0.0725\textwidth}
				\centering
				\includegraphics[width=\linewidth]{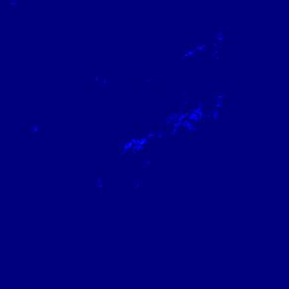}
				\label{fig:sub:p_h_2_5}
			\end{subfigure}
			\begin{subfigure}{0.0725\textwidth}
				\centering
				\includegraphics[width=\linewidth]{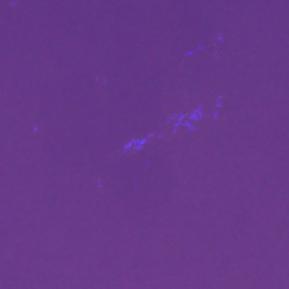}
				\label{fig:sub:p_o_2_5}
			\end{subfigure}
			\hspace{\reshspace}
			\begin{subfigure}{0.0725\textwidth}
				\centering
				\includegraphics[width=\linewidth]{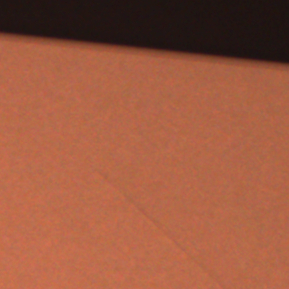}
				\label{fig:sub:p_b_2_6}
			\end{subfigure}
			\begin{subfigure}{0.0725\textwidth}
				\centering
				\includegraphics[width=\linewidth]{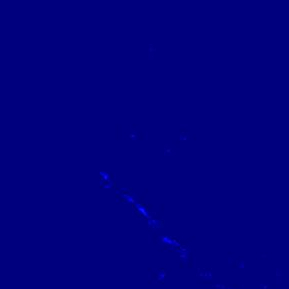}
				\label{fig:sub:p_h_2_6}
			\end{subfigure}
			\begin{subfigure}{0.0725\textwidth}
				\centering
				\includegraphics[width=\linewidth]{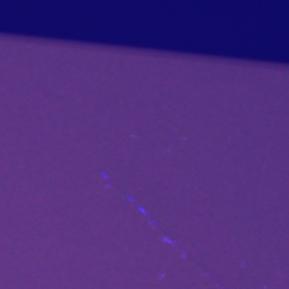}
				\label{fig:sub:p_o_2_6}
			\end{subfigure}
			\vspace{-0.35cm}
		\end{minipage}
	\end{subfigure}
	\begin{subfigure}{\textwidth}
		\begin{minipage}{\indexwidth}
			\centering
			K
			\label{fig:sub:K}
		\end{minipage}
		\begin{minipage}{\inclwidth}
			\begin{subfigure}{0.0725\textwidth}
				\centering
				\includegraphics[width=\linewidth]{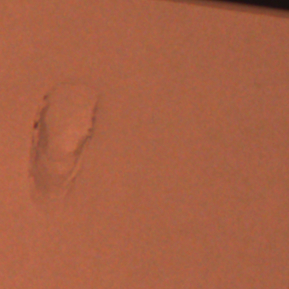}
				\label{fig:sub:p_b_3_1}
			\end{subfigure}
			\begin{subfigure}{0.0725\textwidth}
				\centering
				\includegraphics[width=\linewidth]{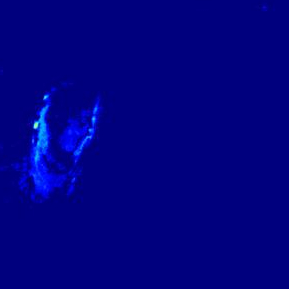}
				\label{fig:sub:p_h_3_1}
			\end{subfigure}
			\begin{subfigure}{0.0725\textwidth}
				\centering
				\includegraphics[width=\linewidth]{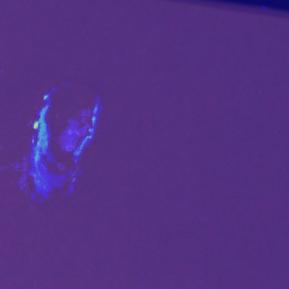}
				\label{fig:sub:p_o_3_1}
			\end{subfigure}
			\hspace{\reshspace}
			\begin{subfigure}{0.0725\textwidth}
				\centering
				\includegraphics[width=\linewidth]{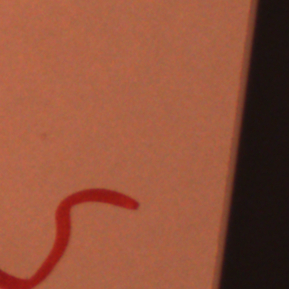}
				\label{fig:sub:p_b_3_2}
			\end{subfigure}
			\begin{subfigure}{0.0725\textwidth}
				\centering
				\includegraphics[width=\linewidth]{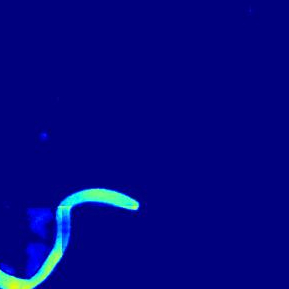}
				\label{fig:sub:p_h_3_2}
			\end{subfigure}
			\begin{subfigure}{0.0725\textwidth}
				\centering
				\includegraphics[width=\linewidth]{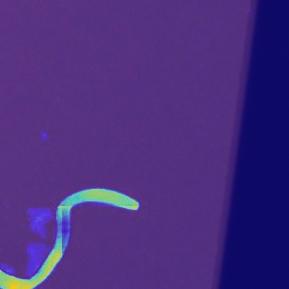}
				\label{fig:sub:p_o_3_2}
			\end{subfigure}
			\hspace{\reshspace}
			\begin{subfigure}{0.0725\textwidth}
				\centering
				\includegraphics[width=\linewidth]{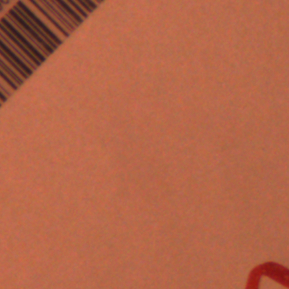}
				\label{fig:sub:p_b_3_3}
			\end{subfigure}
			\begin{subfigure}{0.0725\textwidth}
				\centering
				\includegraphics[width=\linewidth]{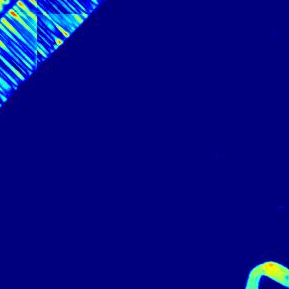}
				\label{fig:sub:p_h_3_3}
			\end{subfigure}
			\begin{subfigure}{0.0725\textwidth}
				\centering
				\includegraphics[width=\linewidth]{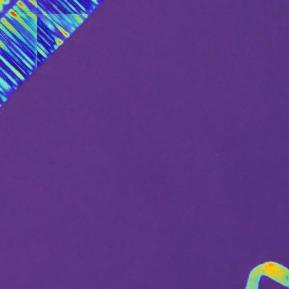}
				\label{fig:sub:p_o_3_3}
			\end{subfigure}
			\hspace{\reshspace}
			\begin{subfigure}{0.0725\textwidth}
				\centering
				\includegraphics[width=\linewidth]{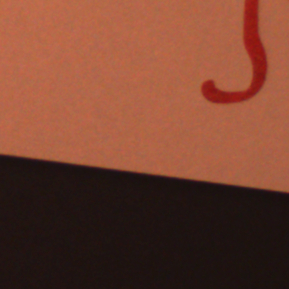}
				\label{fig:sub:p_b_3_4}
			\end{subfigure}
			\begin{subfigure}{0.0725\textwidth}
				\centering
				\includegraphics[width=\linewidth]{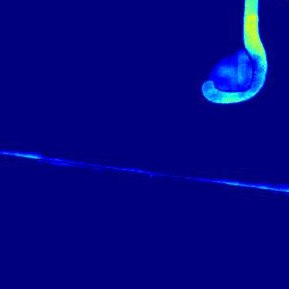}
				\label{fig:sub:p_h_3_4}
			\end{subfigure}
			\begin{subfigure}{0.0725\textwidth}
				\centering
				\includegraphics[width=\linewidth]{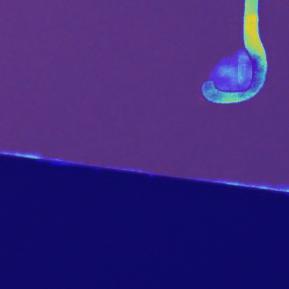}
				\label{fig:sub:p_o_3_4}
			\end{subfigure}
			\vspace{-0.35cm}
		\end{minipage}
	\end{subfigure}
	\begin{subfigure}{\textwidth}
		\begin{minipage}{\indexwidth}
			\centering
			L
			\label{fig:sub:L}
		\end{minipage}
		\begin{minipage}{\inclwidth}
			\begin{subfigure}{0.0725\textwidth}
				\centering
				\includegraphics[width=\linewidth]{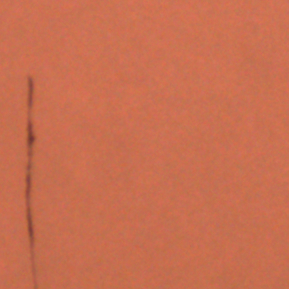}
				\label{fig:sub:p_b_3_5}
			\end{subfigure}
			\begin{subfigure}{0.0725\textwidth}
				\centering
				\includegraphics[width=\linewidth]{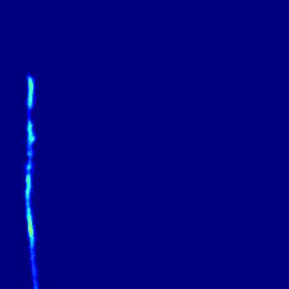}
				\label{fig:sub:p_h_3_5}
			\end{subfigure}
			\begin{subfigure}{0.0725\textwidth}
				\centering
				\includegraphics[width=\linewidth]{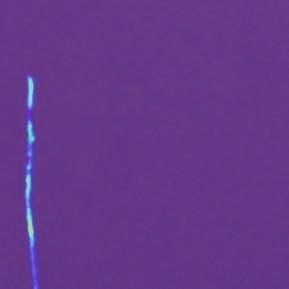}
				\label{fig:sub:p_o_3_5}
			\end{subfigure}
			\hspace{\reshspace}
			\begin{subfigure}{0.0725\textwidth}
				\centering
				\includegraphics[width=\linewidth]{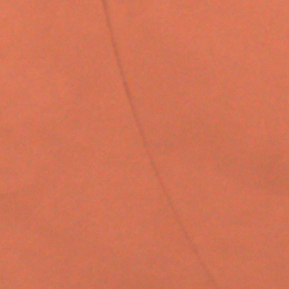}
				\label{fig:sub:p_b_4_2}
			\end{subfigure}
			\begin{subfigure}{0.0725\textwidth}
				\centering
				\includegraphics[width=\linewidth]{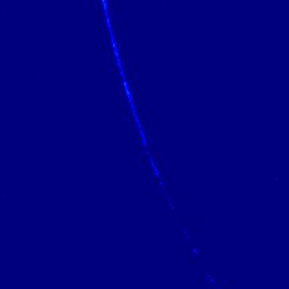}
				\label{fig:sub:p_h_4_2}
			\end{subfigure}
			\begin{subfigure}{0.0725\textwidth}
				\centering
				\includegraphics[width=\linewidth]{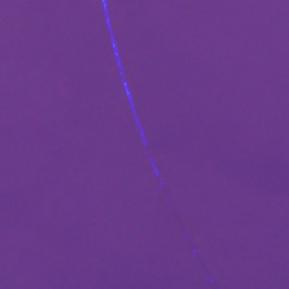}
				\label{fig:sub:p_o_4_2}
			\end{subfigure}
			\hspace{\reshspace}
			\begin{subfigure}{0.0725\textwidth}
				\centering
				\includegraphics[width=\linewidth]{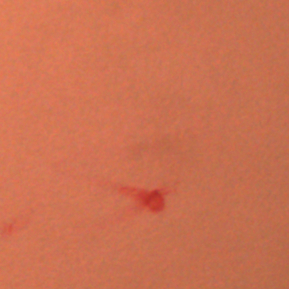}
				\label{fig:sub:p_b_4_5}
			\end{subfigure}
			\begin{subfigure}{0.0725\textwidth}
				\centering
				\includegraphics[width=\linewidth]{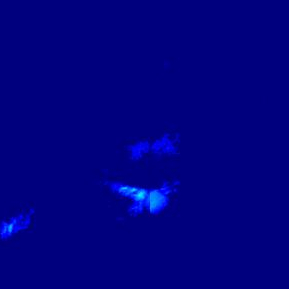}
				\label{fig:sub:p_h_4_5}
			\end{subfigure}
			\begin{subfigure}{0.0725\textwidth}
				\centering
				\includegraphics[width=\linewidth]{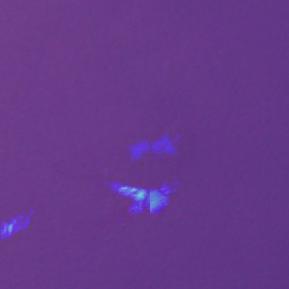}
				\label{fig:sub:p_o_4_5}
			\end{subfigure}
			\hspace{\reshspace}
			\begin{subfigure}{0.0725\textwidth}
				\centering
				\includegraphics[width=\linewidth]{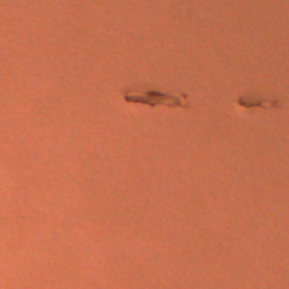}
				\label{fig:sub:p_b_4_6}
			\end{subfigure}
			\begin{subfigure}{0.0725\textwidth}
				\centering
				\includegraphics[width=\linewidth]{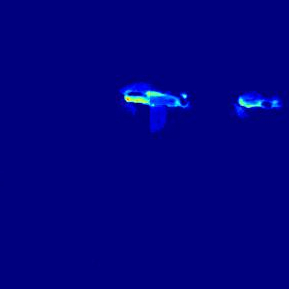}
				\label{fig:sub:p_h_4_6}
			\end{subfigure}
			\begin{subfigure}{0.0725\textwidth}
				\centering
				\includegraphics[width=\linewidth]{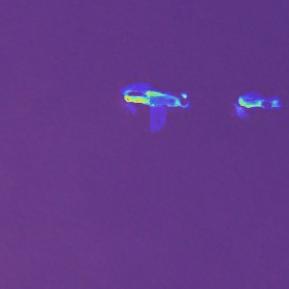}
				\label{fig:sub:p_o_4_6}
			\end{subfigure}
			\vspace{-0.35cm}
		\end{minipage}
	\end{subfigure}
	\begin{subfigure}{\textwidth}
		\begin{minipage}{\indexwidth}
			\centering
			M
			\label{fig:sub:M}
		\end{minipage}
		\begin{minipage}{\inclwidth}
			\begin{subfigure}{0.0725\textwidth}
				\centering
				\includegraphics[width=\linewidth]{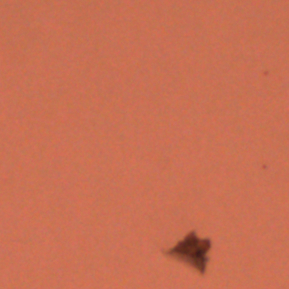}
				\label{fig:sub:p_b_5_10}
			\end{subfigure}
			\begin{subfigure}{0.0725\textwidth}
				\centering
				\includegraphics[width=\linewidth]{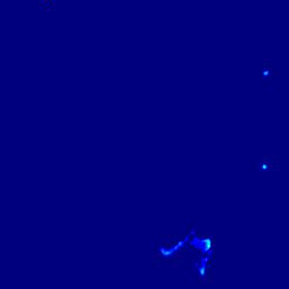}
				\label{fig:sub:p_h_5_10}
			\end{subfigure}
			\begin{subfigure}{0.0725\textwidth}
				\centering
				\includegraphics[width=\linewidth]{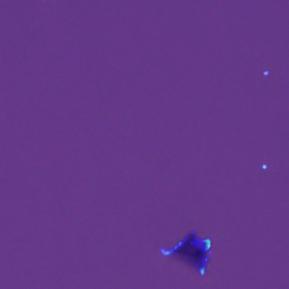}
				\label{fig:sub:p_o_5_10}
			\end{subfigure}
			\hspace{\reshspace}
			\begin{subfigure}{0.0725\textwidth}
				\centering
				\includegraphics[width=\linewidth]{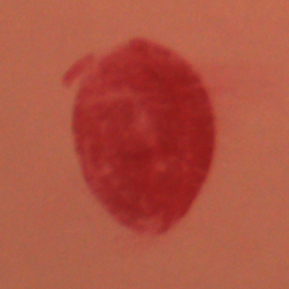}
				\label{fig:sub:p_b_5_20}
			\end{subfigure}
			\begin{subfigure}{0.0725\textwidth}
				\centering
				\includegraphics[width=\linewidth]{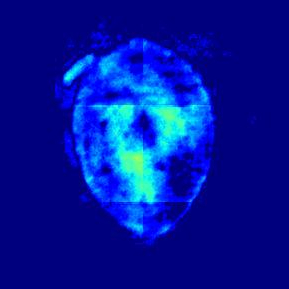}
				\label{fig:sub:p_h_5_20}
			\end{subfigure}
			\begin{subfigure}{0.0725\textwidth}
				\centering
				\includegraphics[width=\linewidth]{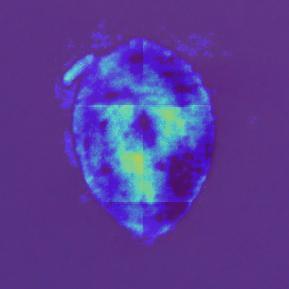}
				\label{fig:sub:p_o_5_20}
			\end{subfigure}
			\hspace{\reshspace}
			\begin{subfigure}{0.0725\textwidth}
				\centering
				\includegraphics[width=\linewidth]{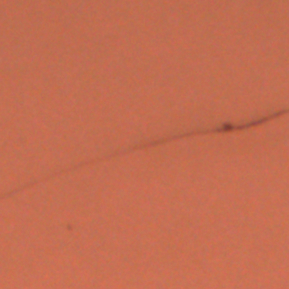}
				\label{fig:sub:p_b_6_3}
			\end{subfigure}
			\begin{subfigure}{0.0725\textwidth}
				\centering
				\includegraphics[width=\linewidth]{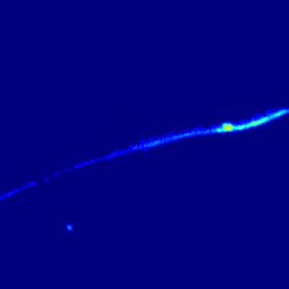}
				\label{fig:sub:p_h_6_3}
			\end{subfigure}
			\begin{subfigure}{0.0725\textwidth}
				\centering
				\includegraphics[width=\linewidth]{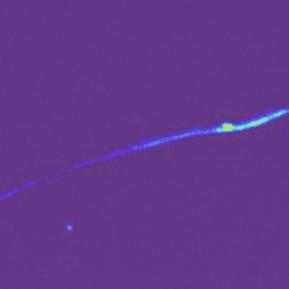}
				\label{fig:sub:p_o_6_3}
			\end{subfigure}
			\hspace{\reshspace}
			\begin{subfigure}{0.0725\textwidth}
				\centering
				\includegraphics[width=\linewidth]{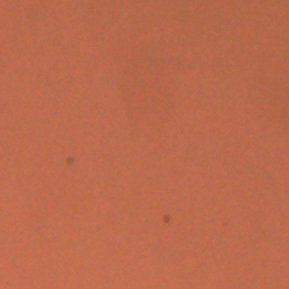}
				\label{fig:sub:p_b_6_4}
			\end{subfigure}
			\begin{subfigure}{0.0725\textwidth}
				\centering
				\includegraphics[width=\linewidth]{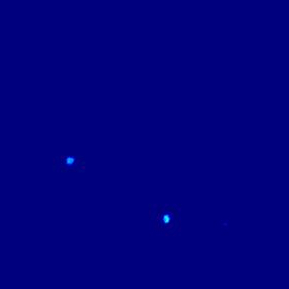}
				\label{fig:sub:p_h_6_4}
			\end{subfigure}
			\begin{subfigure}{0.0725\textwidth}
				\centering
				\includegraphics[width=\linewidth]{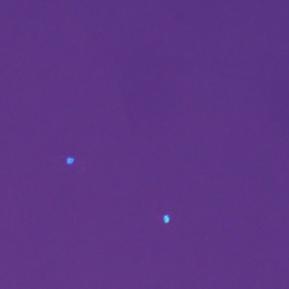}
				\label{fig:sub:p_o_6_4}
			\end{subfigure}
			\vspace{-0.35cm}
		\end{minipage}
	\end{subfigure}
	\begin{subfigure}{\textwidth}
		\begin{minipage}{\indexwidth}
			\centering
			N
			\label{fig:sub:N}
		\end{minipage}
		\begin{minipage}{\inclwidth}
			\begin{subfigure}{0.0725\textwidth}
				\centering
				\includegraphics[width=\linewidth]{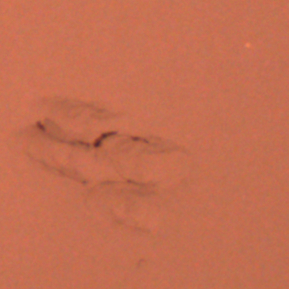}
				\label{fig:sub:p_b_6_12}
			\end{subfigure}
			\begin{subfigure}{0.0725\textwidth}
				\centering
				\includegraphics[width=\linewidth]{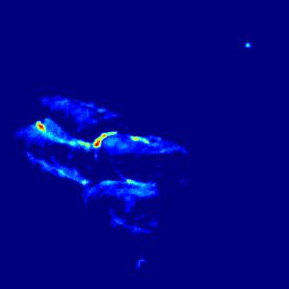}
				\label{fig:sub:p_h_6_12}
			\end{subfigure}
			\begin{subfigure}{0.0725\textwidth}
				\centering
				\includegraphics[width=\linewidth]{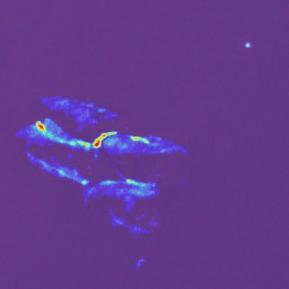}
				\label{fig:sub:p_o_6_12}
			\end{subfigure}
			\hspace{\reshspace}
			\vspace{-0.35cm}
		\end{minipage}
	\end{subfigure}
	\caption{The defect localization on the melamine-faced board involves four columns with the following three grouping sets: the actual anomaly, heatmap of prediction, and the overlay of the heatmap on the original defect, for evaluation. The rows are categorized as follows: corners -> \hyperref[fig:sub:A]{A}, edges -> \hyperref[fig:sub:A]{B - C}, grooves -> \hyperref[fig:sub:A]{D - E}, holes -> \hyperref[fig:sub:A]{F - H}, and plain surface -> \hyperref[fig:sub:A]{I - N}.}
	\label{fig:results_all}
\end{figure}

\section{\MakeUppercase{Conclusion}}
Our paper presents a hybrid approach for detecting surface defects on melamine-faced boards. Unlike traditional methods that use images with a specific region of interest, our methodology utilizes a dataset with images of high-resolution captured with a fixed field-of-view camera, enabling inspection of boards of varying sizes. This model combines techniques, including slicing high-resolution images, addressing imbalanced datasets, performing feature extraction and k-means clustering due to feature frequency variations, and utilizing an autoencoder model for anomaly prediction.

The unsupervised defect localizing model evaluation reveals strong performance in identifying anomalies within the melamine-faced board. The model recognizes and localizes minor artifacts such as smudges, dirt, and deformities, even in corners and edges. It also accurately predicts significant defects around edges and holes, showcasing its robust anomaly detection capabilities. However, challenges arise when dealing with relatively small and blurred anomalies, particularly around the edges. Moreover, the model struggles to identify and localize larger missing sections on melamine-faced boards. The model's behavior around grooves is a mixed bag; it identifies instances of discontinuity but faces difficulties distinguishing artifacts that blend with the surrounding environment. When it comes to plain surface defects, the model performs effectively. It successfully detects large and small surface defects, highlighting its versatility in identifying diverse anomaly types, including those resembling holes in both shape and texture.

In conclusion, the model exhibits promising potential for improving quality control and inspection processes on surfaces of melamine-faced board. This potential is especially evident in its ability to identify defects on plain surfaces, corners, edges, and holes. However, it is necessary to acknowledge the need for continuous refinement to enhance its performance in identifying specific defect types and managing anomalies with less distinct attributes, as discussed in the following subsection.

\subsection{\MakeUppercase{Future scope}}
This paper introduces an approach to localize anomalies on melamine-faced boards, addressing class imbalance challenges while maintaining quality standards. However, let us look at avenues for enhancing our model and its performance in anomaly localization on melamine-faced boards. One potential improvement involves using a higher-resolution camera with a larger aperture to aid in capturing anomalies more effectively, particularly significant ones. Moreover, such a camera could facilitate the imaging of larger boards, contributing to better feature learning with higher memory use. Another approach is the implementation of a camera array, requiring proper calibration and considerations for overlap regions, rectifying barrel distortion to minimize variations in class features, effectively reducing the number of training images. Addressing class imbalance warrants further investigation, especially in identifying damaged corners. One possible strategy is to train separate models for each feature following an initial classifier, involving weighing trade-offs related to the overall model size. Another is to assign weights to each class, which requires an excellent classifier at the initial stage. Optimizing the hyperparameters for Felzenszwalb's segmentation could lead to an artificial dataset representing natural anomalies. This optimization can enhance the model's robustness and improve performance metrics. Exploring alternative architectures, such as UNet++ introduced by \citet{zhou2020unet++}, may offer improved power and generalization capabilities, even though it could lead to additional training parameters. Incorporating techniques like the weighted frequency domain loss, as explored by \citet{nakanishi2021anomaly}, and NST loss, as investigated by \citet{hida2021smart}, could enhance the model's ability to generalize on the edges and textures of board classes.

By pursuing these avenues of improvement, our model could achieve higher accuracy, greater robustness, and improved generalization for localizing anomalies on melamine-faced boards, contributing to elevating quality control standards within the manufacturing industry.

\vspace{6pt} 




\authorcontributions{Conceptualization, D.Mehta; methodology, D.Mehta; software, D.Mehta; formal analysis, D.Mehta; investigation, D.Mehta; data curation, D.Mehta; writing---original draft preparation, D.Mehta.; writing---review and editing, N.Klarmann.; supervision, N.Klarmann.; project administration, N.Klarmann}

\funding{This research received no external funding.}

\conflictsofinterest{The authors declare no conflict of interest. The funders had no role in the design of the study; in the collection, analyses, or interpretation of data; in the writing of the manuscript; or in the decision to publish the results.}



\abbreviations{Abbreviations}{
The following abbreviations are used in this manuscript:\\

\noindent 
\begin{tabular}{@{}ll}
	Adam & Adaptive moment optimization\\
	AUC & Area under curve\\
	AUROC & Area under the receiver operator characteristic\\
	ADM & Artificial defect module\\
	AI & Artificial intelligence\\
	AGV & Automated guided vehicle\\
	CV & Computer vision\\
    DL & Deep learning\\
	DTD & Describable textures dataset\\
	DAE & Denoising autoencoder\\
	FPR & False positive rate\\
	GAN & Generative adversarial network\\
	MSE & Mean square error\\
	NST & Neural style transfer\\
	OpenCV & Open source computer vision library\\
	ROC & Receiver operator characteristics\\
	SSAE & Semisupervised autoencoder\\
	SSIM & Structural similarity\\
	TPR & True positive rate\\
	VQ-VAE & Vector-quantized variational autoencoder
\end{tabular}
}
\begin{adjustwidth}{-\extralength}{0cm}

\reftitle{References}


\externalbibliography{yes}
\bibliography{Bibliogrphy.bib}

%


\PublishersNote{}
\end{adjustwidth}
\end{document}